\documentclass{article}

\usepackage[sort,comma,authoryear,round]{natbib}

\usepackage[preprint]{neurips_2021}

\makeatletter
\def\@fnsymbol#1{\ensuremath{\ifcase#1\or \dagger\or \ddagger\or
   \mathsection\or \mathparagraph\or \|\or **\or \dagger\dagger
   \or \ddagger\ddagger \else\@ctrerr\fi}}
\makeatother

\makeatletter 
\newcommand*{\rom}[1]{\expandafter\@slowromancap\romannumeral #1@}
\makeatother
\newcommand{\RNum}[1]{\uppercase\expandafter{\romannumeral #1\relax}}

\usepackage[utf8]{inputenc} 
\usepackage[T1]{fontenc}    
\usepackage{hyperref}       
\usepackage{url}            
\usepackage{booktabs}       
\usepackage{amsfonts}       
\usepackage{nicefrac}       
\usepackage{microtype}      
\usepackage{xcolor}         
\hypersetup{
    colorlinks=true,
    linkcolor=blue,
    filecolor=magenta,
    citecolor=blue,
    urlcolor=cyan  
}

\hypersetup{
    colorlinks=true,
    linkcolor=blue,
    filecolor=magenta,
    citecolor=blue,
    urlcolor=cyan,
}

\usepackage{appendix}
\usepackage[ruled, vlined]{algorithm2e}  

\usepackage{aliascnt}
\usepackage{hyperref}
\newtheorem{theorem}{Theorem}
\newaliascnt{lemma}{theorem}

\aliascntresetthe{lemma}

\usepackage{amsmath}
\numberwithin{equation}{section}  

\newcommand{\bY}{\mathbf{Y}}
\newcommand{\bU}{\mathbf{U}}
\newcommand{\bV}{\mathbf{V}}
\newcommand{\bW}{\mathbf{W}}
\newcommand{\bH}{\mathbf{H}}
\newcommand{\bS}{\mathbf{S}}
\newcommand{\bA}{\mathbf{A}}

\newcommand{\bu}{\mathbf{u}}
\newcommand{\bv}{\mathbf{v}}
\newcommand{\bw}{\mathbf{w}}
\newcommand{\bx}{\mathbf{x}}
\newcommand{\by}{\mathbf{y}}

\newcommand{\ba}{\mathbf{a}}

\newcommand{\bmeta}{\boldsymbol{\eta}}
\newcommand{\cE}{\mathcal{E}}
\newcommand{\cF}{\mathcal{F}}
\newcommand{\cP}{\mathcal{P}}

\DeclareMathOperator*{\argmin}{arg\,min}
\DeclareMathOperator*{\sign}{sign}

\usepackage{graphicx}
\usepackage{subfig}

\usepackage[english]{babel}
\usepackage{hyperref}
\addto\extrasenglish{
}

\usepackage{verbatim}
\usepackage{ragged2e}
\usepackage{multirow}       
\usepackage{circledsteps}

\usepackage{extarrows}
\usepackage{tikz}
\usetikzlibrary{matrix}
\usepackage{aliascnt}
\usepackage{float}

\title{Robust Matrix Factorization with Grouping Effect}

\author{
    Haiyan Jiang$^1$,
    \quad Shuyu Li$^2$\thanks{The work was done when the author was an intern at Baidu Research.},
    \quad Luwei Zhang$^2$\footnotemark[1],
    \quad Haoyi Xiong$^1$,
    \quad Dejing Dou$^1$ \\
    \\
    $^1$ 
    Baidu Research, Baidu Inc., China \\
    $^2$ Columbia University, New York, NY, USA \\
    \\
    \texttt{jianghaiyan01@baidu.com}, \\
    \texttt{shuryli@outlook.com},
    \texttt{helenzhang0322@gmail.com}, \\
    \texttt{xionghaoyi@baidu.com}, 
    \texttt{doudejing@baidu.com}
}

\begin{document}

\maketitle

\begin{abstract}

Although many techniques have been applied to matrix factorization (MF), they may not fully exploit the feature structure. 
In this paper, we incorporate the grouping effect into MF and propose a novel method called \emph{Robust Matrix Factorization with Grouping effect} (GRMF).
The grouping effect is a generalization of the sparsity effect, which conducts denoising by clustering similar values around multiple centers instead of just around 0.
Compared with existing algorithms, the proposed GRMF can automatically learn the grouping structure and sparsity in MF without prior knowledge, 
by introducing a naturally adjustable non-convex regularization to achieve simultaneous sparsity and grouping effect.
Specifically, GRMF uses an efficient alternating minimization framework to perform MF, in which the original non-convex problem is first converted into a convex problem through Difference-of-Convex (DC) programming, and then solved by Alternating Direction Method of Multipliers (ADMM).
In addition, GRMF can be easily extended to the Non-negative Matrix Factorization (NMF) settings.
Extensive experiments have been conducted using real-world data sets with outliers and contaminated noise, where the experimental results show that GRMF has promoted performance and robustness, compared to five benchmark algorithms.

\textbf{Keywords: } Robust Matrix Factorization, Feature Grouping, Sparsity, Non-convex Regularization

\end{abstract}

\section{Introduction}

Matrix factorization (MF) is a fundamental yet widely-used technique in machine learning~\citep{srebro2005maximum,mnih2007probabilistic,choi2008algorithms,li2019robust,dai2020robust,parvin2019scalable, wang2020robust,chi2019nonconvex}, 
with numerous applications in computer vision~\citep{wang2012probabilistic,wang2013bayesian,gao2022human,haeffele2019structured,xu2020bayesian}, 
recommender systems~\citep{koren2009matrix,xue2017deep,jakomin2020simultaneous}, bioinformatics~\citep{pascual2006bionmf,gaujoux2010flexible,jamali2020mdipa,cui2019rcmf}, social networks~\citep{ma2008sorec,gurini2018temporal,zhang2020graphs}, and many others.
In MF, with a given data matrix $\bY \in \mathbb{R}^{d\times n}$, one is interested in approximating $\bY$ by $\bU \bV^T$ such that the reconstruction error between the given matrix and its factorization is minimized, where $\bU \in \mathbb{R}^{d\times r}, \bV \in \mathbb{R}^{n\times r}$, and $r \ll \min(d, n)$ is the rank. 
Various regularizers have been introduced to generalize the low-rank MF under different conditions and constraints in the following form
\begin{equation}\label{eq:MF-obj}
    \min_{\bU\in \mathbb{R}^{d\times r}, \bV\in \mathbb{R}^{n\times r}}
    \frac{1}{2}\|\bY -\bU\bV^T\|_{\alpha}^{\alpha} + \mathcal{R}_{1}(\bU) + \mathcal{R}_{2}(\bV), 
\end{equation}
where $\mathcal{R}_{1}$ and $\mathcal{R}_{2}$ refer to the regularization terms and $\alpha$ corresponds to the loss types (e.g., $\alpha=2$ for the quadratic loss) for reconstruction errors.
Existing studies~\citep{kim2012group,lin2017robust,abdolali2021simplex} show that the use of regularization terms could prevent overfitting or impose sparsity, while the choice of loss types (e.g., $\alpha=1$ for $\ell_1$-loss) helps to establish robustness of MF with outliers and noise.





\textbf{Background}\quad
The goal of matrix factorization is to obtain the low-dimensional structures of the data matrix while preserving the major information, where singular value decomposition (SVD)~\citep{klema1980singular} and principal component analysis (PCA)~\citep{wold1987principal} are two conventional solutions. To better deal with sparsity in high-dimensional data, 
the truncated-SVD~\citep{hansen1987truncatedsvd} is proposed to achieve result with a determined rank. Like traditional SVD, it uses the $\ell_2$ loss for reconstruction errors, but has additional requirement on the norm of the solution.

\textbf{Loss terms in MF}\quad While the above models could partially meet the goal of MF, they might be vulnerable to outliers and noise and lack robustness. Thus they often give unsatisfactory results on real world data due to the use of inappropriate loss terms for reconstruction errors. 
To solve the problem,~\citet{wang2012probabilistic} proposed a probabilistic method for robust matrix factorization (PRMF) based on the $\ell_1$ loss which is formulated using the Laplace error assumption and optimized with a parallelizable Expectation-Maximization (EM) algorithm. 
In addition to traditional $\ell_1$ or $\ell_2$-loss, researchers have also made attempts with non-convex loss functions.~\citet{yao2017scalable} constructed a scalable robust matrix factorization with non-convex loss, using general non-convex loss functions to make the MF model applicable in more situations. On the contrary,~\citet{haeffele2014structured} employed the non-convex projective tensor norm. 
Yet another algorithm with identifiability guarantee has been proposed in~\citet{abdolali2021simplex}, where the simplex-structured matrix factorization (SSMF) is invented to constrain the matrix $\bV$ to belong to the unit simplex.

\textbf{Sparsity control in MF}\quad In addition to the loss terms, sparsity regularizers or constraints have been proposed to control the number of nonzero elements in MF (either the recovered matrix or the noise matrix). For example,~\citet{hoyer2004non} employed a sparseness measure based on the relationship between the $\ell_1$-norm and the $\ell_2$-norm.~\citet{candes2011robust} introduced a Robust PCA (RPCA) which decomposes the matrix $M$ by $M = L_0+S_0$, where $L_0$ is assumed to have low-rank and $S_0$ is constrained to be sparse by an $\ell_1$ regularizer. 
The Robust Non-negative Matrix Factorization (RNMF)~\citep{zhang2011robust} decomposes the non-negative data matrix as the summation of one sparse error matrix and the product of two non-negative matrices, where the sparsity of the error matrix is constrained with $\ell_1$-norm penalty. 
Compared with the $\ell_2$-norm penalty and the $\ell_1$-norm penalty used in the mentioned methods, the $\ell_0$-norm penalty directly addresses the sparsity measure. 
However, the direct use of the $\ell_0$-norm penalty gives us a non-convex NP-hard optimization problem. To address that difficulty, researchers have proposed different approximations, such as the truncated $\ell_1$ penalty proposed in~\citet{shen2012likelihood}.

\textbf{Our work}\quad
Previous studies on MF pay great attention to sparsity, yet few of them~\citep{li2013cgmf, yuen2012taskrec, ortega2016recommending, rahimpour2017non} take care of the grouping effect. 
While sparse control can only identify factors whose weights are clustered around 0 and eliminate the noise caused by them, the grouping effect can identify factor groups whose weights are clustered around any similar value and eliminate the noise caused by all these groups. 
In fact, the grouping effect can bring intepretability and recovery accuracy to MFs to a larger extent than sparsity alone does. With the desire to conduct matrix factorization with robustness, sparsity, and grouping effect, we construct a new matrix factorization model named \emph{Robust Matrix Factorization with Grouping effect} (GRMF) and evaluate its performance through experiments. 
%

To the best of our knowledge, this work has made unique contributions in introducing sparsity and grouping effect into MF. 
The most relevant works are~\citet{yang2012feature,kim2012group}. Specifically,~\citet{yang2012feature,shen2012likelihood} proposed to use truncated $\ell_1$ penalty to pursue both sparsity and grouping effect in estimation of linear regression models, while the way of using such techniques to improve MF is not known. On the other hand,~\citet{kim2012group} also modeled MF with groups but assumes features in the same group sharing the same sparsity pattern with predefined group partition. Thus a mixed norm regularization is proposed to promote group sparsity in the factor matrices of non-negative MF. Compared to~\citet{kim2012group}, our proposed GRMF is more general-purpose, where the sparsity (e.g., elements in the factor matrix centered around zero) is only one special group in all possible groups for elements in factor matrices.

Specifically, to achieve robustness, the focus of GRMF is on matrix factorization under the $\ell_1$-loss.
GRMF further adopts an $\ell_0$ surrogate---truncated $\ell_1$, for the penalty term to control the sparsity and the grouping effect.
As the resulting problem is non-convex, we solve it with difference-of-convex algorithm (DC) and Alternating Direction Method of Multipliers (ADMM). The algorithms are implemented to conduct experiments on COIL-20, ORL, extended Yale B, and JAFFE datasets. In addition, we compare the result with 5 benchmark algorithms: 
Truncated SVD~\citep{hansen1987truncatedsvd},
RPCA~\citep{candes2011robust}, RNMF~\citep{wen2018survey},
RPMF~\citep{wang2012probabilistic} and  GoDec+~\citep{guo2017godec+}. The results show that the proposed GRMF significantly outperforms existing benchmark algorithms.

The remainder of the paper is organized as follows. In Section 2, we briefly introduce the robust MF and the non-negative MF. We propose the GRMF in Section 3 and give the algorithm for GRMF which uses DC and ADMM for solving the resulting non-convex optimization problem in Section 4. Experiment results to show the performances of GRMF and comparison with other benchmark algorithms are presented in Section 5. We conclude the paper in Section 6.

\textbf{Notations}\quad
For a scalar $x$, $\sign(x)=1$ if $x>0$, $0$ if $x=0$, and $-1$ otherwise. 
For a matrix $\bA$, $ { \|\bA\|_F=(\sum_{i,j} A_{ij}^2)^{\frac{1}{2}} } $ is its Frobenius norm, and $\|\bA\|_{\ell_1}= \sum_{i,j} |A_{ij}| $ is its $\ell_1$-norm. 
For a vector $\bx$, $\|\bx\|_{\ell_1}= \sum_{i} |x_{i}| $ is its $\ell_1$-norm.
Denote data matrix $ { \bY_{d\times n} = [\by_{\cdot 1}, \cdots, \by_{\cdot n}] } $ the stack of $n$ column vectors with each $\by_{\cdot j} \in \mathbb{R}^{d}$, or $ { \bY^T = [\by_{1\cdot}^T, \cdots, \by_{d\cdot}^T] } $ the stack of $d$ row vectors with each $\by_{i\cdot}\in \mathbb{R}^{n}$.
Denote ${ \bU^T = [\bu_1, \cdots, \bu_d] \in \mathbf{R}^{r\times d} } $, and ${ \bV^T=[\bv_1, \cdots, \bv_n] \in \mathbf{R}^{r\times n} } $, with each $ {{\bu_1, \cdots, \bu_d, \bv_1, \cdots, \bv_n} \in \mathbb{R}^{r} } $. 
Denote $\bv_{jl}$ the $l$-th element of the vector $\bv_j \in \mathbb{R}^{r}$ and $\bu_{il}$ the $l$-th element of the vector $\bu_i \in \mathbb{R}^{r}$.

\section{Preliminaries }
\paragraph{Robust matrix factorization (RMF)}

Given a data matrix $\bY \in \mathbb{R}^{d\times n}$, the matrix factorization~\citep{yao2017scalable} is formulated as $\bY = \bU\bV^T + \boldsymbol{\varepsilon} $. 
Here $\bU \in \mathbb{R}^{d\times r}, \bV \in \mathbb{R}^{n\times r}$, $r \ll \min(d, n)$ is the rank, and $\boldsymbol{\varepsilon}$ is the noise/error term. 
The RMF can be formulated under the Laplacian error assumption, $\min_{\bU, \bV} \|\bY - \bU\bV^T\|_{\ell_1} = \sum_{i=1}^{d}\sum_{j=1}^{n} |y_{ij} - \bu_i^T\bv_j|$,
where ${ \bU^T = [\bu_1, \cdots, \bu_d] \in \mathbf{R}^{r\times d} } $, and ${ \bV^T=[\bv_1, \cdots, \bv_n] \in \mathbf{R}^{r\times n} } $. Adding regularizers gives the optimization problem,
$ { \min_{\bU \in \mathbb{R}^{d\times r}, \bV \in \mathbb{R}^{n\times r}}
  \| \bY - \bU\bV^{T}\|_{\ell_1} + \frac{\lambda_u}{2}\|\bU\|_2^2 + \frac{\lambda_v}{2}\|\bV\|_2^2 } $.

\paragraph{Robust non-negative matrix factorization (RNMF)}
As the traditional NMF is optimized under the Gaussian noise or Poisson noise assumption, RNMF~\citep{zhang2011robust} is introduced to deal with data that are grossly corrupted. RNMF decomposes the non-negative data matrix as the summation of one sparse error matrix and the product of two non-negative matrices. The formulation states
$\min_{\bW, \bH, \bS}\|\bY-\bW\bH-\bS\|^2_F + \lambda\|\bS\|_{\ell_1}\; \; \text { s.t. } \bW\geq 0, \bH \geq 0$, 
where $\|\cdot\|_F$ is the Frobenius norm, $\|\bS\|_{\ell_1} = \sum_{i=1}^{d}\sum_{j=1}^{n}|S_{ij}|$, and $\lambda >0 $ is the regularization parameter, controlling the sparsity of $\bS$.

\section{Proposed GRMF formulation}

In this section, we introduce our \emph{Robust Matrix Factorization with Grouping effect} (GRMF) by incorporating the grouping effect into MF.
The problem of GRMF is formulated as follows 
\begin{equation}\label{eq:grmf-obj}
  \min_{\bU \in \mathbb{R}^{d\times r}, \bV \in \mathbb{R}^{n\times r} } f(\bU, \bV) = 
  \|\bY - \bU\bV^{T}\|_{\ell_1} + \mathcal{R}(\bU) + \mathcal{R}(\bV)\ , 
\end{equation}
where $\mathcal{R}(\bU)$ and $\mathcal{R}(\bV)$
are two regularizers corresponding to $\bU$ and $\bV$, given by
\begin{equation*}
\begin{split}
\mathcal{R}(\bU) = &
  \sum_{i=1}^{d} \lambda_{1} \cP_1(\bu_i)
  + \sum_{i=1}^{d} \lambda_{2} \cP_2(\bu_i)
  + \sum_{i=1}^{d} \lambda_{3} \cP_3(\bu_i), 
  \text{ and } \\
  \mathcal{R}(\bV) = &
  \sum_{j=1}^{n} \lambda_{1} \cP_1(\bv_j)
  + \sum_{j=1}^{n} \lambda_{2} \cP_2(\bv_j)
  + \sum_{j=1}^{n} \lambda_{3} \cP_3(\bv_j)\ . 
\end{split}
\end{equation*}
Here $\cP_1(\cdot)$, $\cP_2(\cdot)$ and $\cP_3(\cdot)$ are different penalty functions, which take the following form with respect to a vector $\bx \in \mathbb{R}^{r}$, 
\begin{equation}
\cP_1(\bx) = \sum_{l=1}^{r} \min\left(\frac{|x_l|}{\tau_1}, 1\right), \ 
\cP_2(\bx) = 
 \sum_{l<l': (l, l')\in \cE} \min\left(\frac{|x_{l}-x_{l'}|}{\tau_2}, 1\right), \ 
\cP_3(\bx) = \sum_{l=1}^{r} x_l^2\ ,
\end{equation}
where $\cP_1(\cdot)$ and $\cP_2(\cdot)$ are two regularization terms controlling the sparsity (feature selection) and the grouping effect (feature grouping), $\tau_1$ and $\tau_2$ are thresholding parameters asserting when a small weight or a small difference between two weights will be penalized, $\lambda_1$ and $\lambda_2$ are the corresponding tuning parameters. Here $\cP_3(\cdot)$ is an inherited penalty term from MF with its tuning parameter $\lambda_3$. 
The truncated $\ell_1$-norm penalty $\cP_1(\cdot)$ can be viewed as a surrogate of the $\ell_0$-norm penalty for feature selection~\citep{shen2012likelihood}, where $\min(|x_l|/\tau, 1)$ is an approximation of $I(x_l \neq 0)$ when $\tau \rightarrow 0$.
In addition, GRMF can be extended to non-negative MF (see details in {\color{blue}Appendix B}).
For other notations, denote $\cE=\{(l,l'): l\neq l', l,l'=1,\cdots, r\}$ a set of edges between two distinct nodes $l\neq l'$ of an undirected complete graph.

In the proposed GRMF, we adopt the $\ell_1$-loss to attain the robustness 
and introduce a naturally adjustable non-convex regularization to achieve simultaneous sparsity and grouping effect.
Due to the non-convex regularization and the low-rank constraint in MF, the GRMF is a non-convex problem. By fixing $\bU$ or $\bV$ and updating the other one, GRMF becomes solvable.

\section{Algorithms for GRMF}

The GRMF problem includes the optimization of two matrices $\bU$ and $\bV$, which are treated as two independent variables in the alternative optimization process, while the other is fixed. 


Note that 
\begin{equation*} 
\|\bY - \bU\bV^{T}\|_{\ell_1} 
= \sum_{j=1}^{n} \|\by_{\cdot j} - \bU \bv_j\|_{\ell_1} 
= \sum_{i=1}^{d} \|\by_{i\cdot } - \bV \bu_i \|_{\ell_1}\ .
\end{equation*}
During the alternative optimization procedure, $\bU$ and $\bV$ should be updated alternatively according to $L(\bU|\bV)$ and $L(\bV|\bU)$, which are given by 
\begin{align}
& \min_{\bU \in \mathbb{R}^{d\times r}} L(\bU|\bV) 
  = \sum_{i=1}^{d} \|\by_{i \cdot} - \bV \bu_i\|_{\ell_1}
  + \sum_{i=1}^{d} \lambda_{1} \cP_1(\bu_i)
  + \sum_{i=1}^{d} \lambda_{2} \cP_2(\bu_i)
  + \sum_{i=1}^{d} \lambda_{3} \cP_3(\bu_i), \label{eq:LU-fixV} \\
& \min_{\bV \in \mathbb{R}^{n\times r}} L(\bV|\bU)
  = \sum_{j=1}^{n} \|\by_{\cdot j} - \bU \bv_j\|_{\ell_1} 
  + \sum_{j=1}^{n} \lambda_{1} \cP_1(\bv_j)
  + \sum_{j=1}^{n} \lambda_{2} \cP_2(\bv_j)
  + \sum_{j=1}^{n} \lambda_{3} \cP_3(\bv_j).  \label{eq:LV-fixU}
\end{align}
Note that the optimization problem of $L(\bU|\bV)$ can be decomposed into $d$ independent subproblems. The same procedure applies to the minimization of $L(\bV|\bU)$. 
Thus, the problem of GRMF is a combination of $n+d$ optimization subproblems, each with respect to a vector in $\mathbb{R}^r$. 
For each subproblem, we use the difference-of-convex algorithm (DC) to approximate a non-convex cost function. At each iteration, a quadratic problem with equality constraints is solved by the Alternating Direction Method of Multipliers (ADMM).

The algorithm for GRMF consists of three steps. \emph{First}, we apply an alternative minimization framework to decompose the problem of GRMF into $d+n$ independent subproblems, each optimizing $\bv_j$ (a column in $\bV^T$) or $\bu_i$ (a column in $\bU^T$). \emph{Second}, the non-convex regularization function in each subproblem is decomposed into a difference of two convex functions, and, through linearizing the trailing convex function, a sequence of approximations is constructed by its affine minorization. \emph{Third}, the constructed quadratic problem with equality constraints is solved by ADMM.

\begin{algorithm}[H] 
\SetAlgoLined
\KwIn{\emph{Initialization} $\bU^{(0)}$, $\bV^{(0)}$; tuning parameters $\lambda_1, \lambda_2, \lambda_3, \tau_1, \tau_2$; small tolerance $\delta_{\bU}, \delta_{\bV}$, $t = 1$.}
\KwResult{\emph{Optimal} $\bU, \bV$.}
 \While{ $ \|\bU^{(t)} - \bU^{(t-1)}\|_F^2<\delta_{\bU}$ or 
 $ \|\bV^{(t)} - \bV^{(t-1)}\|_F^2<\delta_{\bV}$ } 
   {
     \For{$j = 1, \cdots, n$}{
     Update $\bv_j^{(t)} = \argmin_{\bv_j \in \mathbb{R}^{r}} L(\bv_j|\bU^{(t-1)})$ by~\autoref{alg:ui-update}\;
     }
     Then update $[\bV^{(t)}]^T =[\bv_1^{(t)}, \cdots, \bv_n^{(t)}]$\;
     \For{$i = 1, \cdots, d$}{
     Update $\bu_i^{(t)} = \argmin_{\bu_i \in \mathbb{R}^{r}} L(\bu_i|\bV^{(t)})$ by~\autoref{alg:ui-update}\;
     }
     Then update $[\bU^{(t)}]^T =[\bu_1^{(t)}, \cdots, \bu_d^{(t)}]$\;
     $t \leftarrow t + 1$\;
 }
\caption{The alternating minimization algorithm for GRMF}
\label{alg:UV-update}
\end{algorithm}

\subsection{Alternative minimization framework for GRMF}

To solve the GRMF problem, we alternatively update $\bV^T=[\bv_1, \cdots, \bv_n]\in \mathbb{R}^{r \times n}$ (while fixing $\bU$) and $\bU^T=[\bu_1, \cdots, \bu_d]\in \mathbb{R}^{r\times d}$ (while fixing $\bV$). 
Thus the alternative minimization framework at the $t$-th step consists of updating ${ \bU^{(t)} = \argmin_{\bU} L(\bU|\bV^{(t)})}$ and ${ \bV^{(t)} = \argmin_{\bV} L(\bV|\bU^{(t-1)})}$ iteratively until convergence.
%
%
%
Problem~\eqref{eq:LU-fixV} is therefore decomposed into $d$ independent small problems, each one optimizing $\bu_i$, ${\bu_i = \argmin_{\bx \in \mathbb{R}^{r}} L(\bx|\bV^{(t)}) }$,
\begin{equation} \label{eq:ui-formula}
\begin{split}
\bu_i = & \argmin_{\bx \in \mathbb{R}^{r}} \Bigg\{ 
  \|\by_{i \cdot } - \bV^{(t)}\bx\|_{\ell_1}
  + \lambda_{1} \sum_{l=1}^{r} \min\left(\frac{|x_l|}{\tau_1}, 1\right)  \\
  & + \lambda_{2}\sum_{l<l': (l,l')\in \cE_{\bV_j}} \min\left(\frac{|x_l-x_{l'}|}{\tau_2}, 1\right)
  + \lambda_{3} \sum_{l=1}^{r} x_l^2 \Bigg\}\ ,
\end{split}
\end{equation}
where $\lambda_{1}$ is a parameter controlling the sparsity and $\lambda_{2}$ controls the grouping effect. Problem~\eqref{eq:LV-fixU} can be decomposed into $n$ independent small subproblems, each optimizing with respect to $\bv_j$, with similar structures.
We propose an alternating minimization framework to solve the GRMF problem (\autoref{alg:UV-update}), by updating $\bU$ and $\bV$ alternatively until convergence.

\begin{theorem}\label{theo:outer-converge}
\emph{(Global Convergence of GRMF).} Let $\bY$ be a matrix in $\mathbb{R}^{d \times n}$. Let $\left\{(\bU^{(t)}, \bV^{(t)})\right\}_{t\in \mathbb{N}}$ be a bounded sequence generated by ~\autoref{alg:UV-update} to solve GRMF, with $\lambda_1, \lambda_2, \lambda_3, \tau_1, \tau_2$ appropriately selected. Then the sequence has finite length and converges to a critical point.
\end{theorem}

As for convergence analysis,~\autoref{theo:outer-converge} gives a convergence guarantee of the alternative minimization framework to solve the GRMF problem. 
The main theorem of~\citet{Csiszar} (\emph{Theorem 3}) proves a general convergence behavior of alternating minimization, and~\citet{hardt2014understanding} (\emph{Theorem 3.8}) proves the convergence of a specific application of alternating minimization algorithm in MF. As a special case of alternating minimization applied in MF, the convergence behavior of GRMF is thus summarized in~\autoref{theo:outer-converge}. The proof just mimics the proof of \emph{Theorem 3.8} in~\citet{hardt2014understanding}, and we omit the details, as~\autoref{alg:UV-update} is a special case of alternating minimization algorithm applied in MF.

\paragraph{Limitation of the model}
One limitation of GRMF is that it uses non-convex regularizers which are neither smooth nor differentiable.~\citet{na2019nonconvex} has pointed out that, when using non-convex penalties, iterative methods such as gradient or coordinate descent may terminate undesirably at a local optimum, which can be different from the global optimum we pursue. As is pointed out in~\citet{wen2018survey},
the performance of non-convex optimization problems is usually closely related to the initialization. These are the inherent drawbacks of non-convex optimization problems.

\subsection{General formulation for each GRMF subproblem}

The minimization problem~\eqref{eq:ui-formula} can be viewed as a special form of the following constrained regression-type optimization problem with simultaneous supervised grouping and feature selection,
\begin{equation}\label{eq:obj-general}
\min_{\bx \in \mathbb{R}^{r}}
 \|\bA\bx - \mathbf{b} \|_{\ell_1}
  + \lambda_1 \sum_{l=1}^{r} \min\left(\frac{|x_l|}{\tau_1}, 1\right)
  + \lambda_2 \sum_{l<l': (l, l')\in \cE} \min\left(\frac{|x_{l}-x_{l'}|}{\tau_2}, 1\right)
  + \lambda_3 \sum_{l=1}^{r} x_l^2 \ .
\end{equation}

The notation $\mathbf{b} \in \mathbb{R}^{n}$ in~\eqref{eq:obj-general} corresponds to $\by_{i \cdot }  \in \mathbb{R}^{n}$ in~\eqref{eq:ui-formula}, and $\bA \in\mathbb{R}^{n\times r}$ corresponds to $\bV^{(t)} \in\mathbb{R}^{n\times r}$. Thus $\|\bA\bx - \mathbf{b} \|_{\ell_1} = \sum_{i=1}^{n} |b_i - \ba_i^T\bx| $, where $\ba_i \in \mathbb{R}^{r}$.
Here $\lambda_1, \lambda_2$ are positive tuning parameters controlling feature sparsity and grouping effect, $\lambda_3>0$ is to prevent overfitting, $\tau_1>0$ is a thresholding parameter determining when a large regression coefficient will be penalized in the model, and $\tau_2>0$ determines when a large difference between two coefficients will be penalized. 
$\mathcal{E} \subset \{1, \cdots, r\}^2$ denotes a set of edges between distinct nodes $l\neq l'$ for a complete undirected graph, with $l\sim l'$ representing an edge directly connecting two nodes. Note that the edge information on the complete undirected graph $\mathcal{E}$ is unknown, and need be learned from the model.

\paragraph{The DC algorithm}
To solve problem~\eqref{eq:obj-general}, considering the non-smooth property of the $\ell_1$ loss, we use a smooth approximation 
$|b_i - \ba_i^T\bx| \approx ((b_i - \ba_i^T\bx)^2 + \epsilon )^{1/2}$, 
where $\epsilon$ is chosen to be a very small positive value, e.g. $10^{-6}$. 
For each alternative minimization, we need to solve the following optimization problem, 
Then optimization problem~\eqref{eq:obj-general} becomes minimizing
\begin{equation*}
S(\bx) = \sum_{i=1}^n ((b_i - \ba_i^T\bx)^2 + \epsilon)^{1/2}
    + \lambda_1 \sum_{l=1}^{r} \min  \left(\frac{|x_l|}{\tau_1}, 1 \right) 
    + \lambda_2 \sum_{l<l':(l,l')\in \mathcal{E}}^{r} \min \left(\frac{|x_l-x_{l'}|}{\tau_2}, 1 \right)
    + \lambda_3\|\bx\|_2^2 \ .
\end{equation*} 
By applying $\min (a,b) = a-(a-b)_{+}$, the two regularizers can be rewritten and $S(\bx)$ can be decomposed into a difference of two convex functions,
$S_{1}(\bx) - S_{2}(\bx)$, which are defined as follows respectively,
\begin{equation*}
  \begin{split}
     S_1(\bx) & = \sum_{i=1}^n ((b_i - \ba_i^T\bx)^2 + \epsilon)^{1/2} 
    + \lambda_1 \sum_{l=1}^{r} \frac{|x_l|}{\tau_1} 
    + \lambda_2 \sum_{l<l':(l,l')\in \mathcal{E}} \frac{|x_l-x_{l'}|}{\tau_2} + \lambda_3 \sum_{l=1}^{r} x_l^2  \\
    S_2(\bx) & = \lambda_1 \sum_{l=1}^{r} \left(\frac{|x_l|}{\tau_1} - 1\right)_{+} 
    + \lambda_2 \sum_{l<l':(l,l')\in \mathcal{E}} \left(\frac{|x_l-x_{l'}|}{\tau_2}-1\right)_{+} \ .
  \end{split}
\end{equation*}
At the $m$-th iteration, $S_{2}(\bx)$ should be linearized as a linear approximation of $S_{2}(\bx)$ at (m-1)-th iteration. Thus, for the $m$-th minimization, we need to solve the following subproblem (see {\color{blue}{Appendix A.1}} for details)
%
%
%
\begin{equation}\label{eq:m-beta}
S^{(m)}(\bx) = 
  \sum_{i=1}^n ((b_i - \ba_i^T\bx)^2 + \epsilon)^{1/2}
  + \frac{\lambda_1}{\tau_1}\sum_{l \in \mathcal{F}^{(m-1)}} |x_l| 
+ \frac{\lambda_2}{\tau_2}\sum_{l<l': (l,l') \in \mathcal{E}^{(m-1)}} |x_l - x_{l'}| 
  + \lambda_3 \|\bx\|_2^2,
\end{equation}
where
\begin{equation}\label{eq:EF-update}
\begin{split}
  \mathcal{F}^{(m-1)} & = \left\{l:|\hat{x}_l^{(m-1)}|<\tau_1 \right\},
  \;\; \\
  \mathcal{E}^{(m-1)} & = \left\{(l,l')\in \mathcal{E}, 
    |\hat{x}_l^{(m-1)} - \hat{x}_{l'}^{(m-1)}|<\tau_2 \right\} \ .
\end{split}
\end{equation}
Here $\hat{\bx}^{(m)} = \argmin_{\bx} 
S^{(m)}(\bx)$, and $\hat{\bx}^{(m-1)}$ is the result at the (m-1)-th iteration. Denote $x_{ll'} = x_l - x_{l'}$, 
and define $ { {\boldsymbol{\xi}} = (x_1, \cdots, x_r, x_{12}, \cdots, x_{1r}, \cdots, x_{(r-1)r}) }$. The $m$-th subproblem~\eqref{eq:m-beta} can be reformulated as an equality-constrained convex optimization problem, 
\begin{equation}\label{eq:constraint}
S^{(m)}(\boldsymbol{\xi}) 
  = \sum_{i=1}^n ((b_i - \ba_i^T \bx)^2 + \epsilon)^{1/2}
    + \frac{\lambda_1}{\tau_1} \sum_{l \in \mathcal{F}^{(m-1)}} |x_l| 
    + \frac{\lambda_2}{\tau_2} \sum_{l<l': (l,l')\in \mathcal{E}^{(m-1)}} | x_{ll'}| 
    + \lambda_3 \|\bx\|_2^2\ ,
\end{equation}
subject to $x_{ll'} = x_l - x_{l'}, \quad  \forall l<l':(l,l')\in \mathcal{E}^{(m-1)}$.

\paragraph{Alternating direction method of multipliers}
In this step, at each iteration, a quadratic problem with linear equality constraints is solved by the augmented Lagrange method. The augmented Lagrangian form of~\eqref{eq:constraint} is as follows (see details in {\color{blue}Appendix A.2})
\begin{equation*}
  L_{\nu}^{(m)}(\boldsymbol{\xi}, \boldsymbol{\tau})
  = S^{(m)} (\boldsymbol{\xi})
  + \sum_{l<l': (l,l')\in \mathcal{E}^{(m-1)}} \tau_{ll'}(x_l - x_{l'} - x_{ll'}) 
   + \frac{\nu}{2} \sum_{l<l': (l,l')\in \mathcal{E}^{(m-1)}} (x_l - x_{l'} - x_{ll'})^2 \ .
\end{equation*}


Here $\tau_{ll'}$ and $\nu$ are the Lagrangian multipliers for the linear constraints and for the computational acceleration. 
Update $\tau_{ll'}$ and $\nu$, with a constant value $\alpha>1$ to accelerate the convergence,
\begin{equation}\label{eq:uv-update}
\tau_{ll'}^{k+1} = \tau_{ll'}^{k} 
+ \nu^{k} (\hat{x}_l^{(m,k)} - \hat{x}_{l'}^{(m,k)} - \hat{x}_{ll'}^{(m,k)}), \quad
\nu^{k+1} = \alpha \nu^k \ .
\end{equation}

To compute ${\widehat{\boldsymbol{\xi}}^{(m,k)} 
 = \argmin_{\boldsymbol{\xi}} L^{(m)}_{\nu} (\boldsymbol{\xi}, \boldsymbol{\tau}^k)} $, 
we use ADMM and alternatively update between $\bx$ and $x_{ll'}$, while fixing ${\tau}_{ll'}^k$ and $\nu^k$ (for $k = 1,2,\cdots$). The following two items give the final results (check the details in {\color{blue}Appendix A.3}). 

\begin{itemize}
    \item 
  Given $\hat{x}_l^{(m,k-1)}$, update $\hat{x}_l^{(m,k)} $ by 
  \begin{equation}\label{eq:xl-update}
     \hat{x}_l^{(m,k)} = \alpha^{-1} \gamma \quad (l = 1,\cdots,r) \ ,
  \end{equation}
  where 
  \[ \alpha = \sum_{i=1}^{n} D_i^{-\frac{1}{2}} A_{il}^2
        + 2\lambda_3  
        + \nu^k \left\vert l':(l,l') \in \mathcal{E}^{(m-1)}\right\vert . \]
  Let
\begin{equation*}
\gamma^{\star} = 
    \sum_{i=1}^n D_i^{-\frac{1}{2}} c_{il}
    - \sum_{l<l': (l,l') \in \mathcal{E}^{(m-1)}} \tau^k_{ll'} 
   + \nu^k \sum_{l<l': (l,l') \in \mathcal{E}^{(m-1)}} (\hat{x}_{l'}^{(m,k-1)} 
    + \hat{x}_{ll'}^{(m,k-1)}) \ .
\end{equation*}

%
%
%
Then $\gamma= \gamma^*$ if $|\hat{x}_l^{(m-1)}| \geq \tau_1$; otherwise $\gamma= \mathrm{ST}(\gamma^*, \frac{\lambda_1}{\tau_1})$. 
$\mathrm{ST}(\cdot, \cdot)$ is the soft threshold function ${ \mathrm{ST}(b, a) = \sign(b) (|b| - a)_{+}}$. Here $ { D_i = (b_i-\ba_i^T \widehat{\bx}^{(m,k-1)})^2+\epsilon }$,
  and $ { c_{il} = A_{il} (b_i -\ba_{i,-l}^T \widehat{\bx}_{-l}^{(m,k-1)}) } $.
  $\ba_{i,-l} $ is the $i$-th row vector of $\bA$ without the $l$-th column, and $\widehat{\bx}_{-l}^{(m,k-1)}$ is the same as $\widehat{\bx}^{(m,k-1)}$ without the $l$-th element. 
  $\epsilon$ is often set to be a small positive constant $10^{-6}$.

\item 
    Given $\hat{x}_{ll'}^{(m,k-1)}$, update $\hat{x}_{ll'}^{(m,k)} $ (with $\hat{x}_l^{(m,k)}$ already updated and fixed).
    \begin{equation}\label{eq:xll-update}
        \hat{x}_{ll'}^{(m,k)} = 
            \begin{cases}
            \frac{1}{\nu^k} \mathrm{ST}\left(\tau_{ll'}^k 
            + \nu^k(\hat{x}_l^{(m,k)} - \hat{x}_{l'}^{(m,k)}), \frac{\lambda_2}{\tau_2} \right), 
            & \text{ if } (l, l') \in \mathcal{E}^{(m-1)}  \\
            \hat{x}_{ll'}^{(m-1)}, 
            & \text{ if }(l, l') \not\in \mathcal{E}^{(m-1)}
            \end{cases} \ . 
    \end{equation}
\end{itemize}

\begin{algorithm} 
\SetAlgoLined
\KwIn{\emph{Initialization $\bx^{(0)} $, tuning parameters $\lambda_1, \lambda_2, \lambda_3, \tau_1, \tau_2 $.} }
\KwResult{$\bu_i^{(t)}$. The resulting $\bx$ is assigned to $\bu_i^{(t)}$. }
 Assign $\cF^{(0)}$, $\cE^{(0)}$ given $\widehat{\bx}^{(0)}$. 
 Assign $\hat{x}_{ll'}^{(0)} = \hat{x}_l^{(0)} - \hat{x}_{l'}^{(0)}, \forall l<l'(l, l') \in \cE^{(0)}$, m=1\;
 $\hat\bx^{(m,0)} = \hat\bx^{(m-1)}$, 
     $\hat{x}_{ll'}^{(m,0)} = \hat{x}_{ll'}^{(m-1)} \;\; \forall\; l<l'\text{ and }(l, l') \in \cE^{(0)}$\;
 \While{$\|\hat{x}_l^{(m)} - \hat{x}_l^{(m-1)}\|^2_2 > \epsilon_{\mathrm{outer}}$}{
     $m \leftarrow m+1$\;
     Update $\cF^{(m)}, \cE^{(m)}$ by~\eqref{eq:EF-update} given $\widehat{\bx}^{(m)}$\;
     $\hat{x}_{ll'}^{(m)} = \hat{x}_l^{(m)} - \hat{x}_{l'}^{(m)}, \forall l<l'\text{ and }(l, l') \in \cE^{(m)} $\;
     \While{$\|\hat{x}_l^{(m, k)} - \hat{x}_l^{(m, k-1)}\|^2_2 > \epsilon_{\mathrm{inner}}$}{
       $k \leftarrow k+1$\;
       Update $\tau_{ll'}^{k}$ and $\nu^{k}$ by~\eqref{eq:uv-update}\;
       Update $\hat{x}_{l}^{(m,k)}$ by~\eqref{eq:xl-update}\;
       Update $\hat{x}_{ll'}^{(m,k)}$ by~\eqref{eq:xll-update}\;
     }
 }
\caption{The DC-ADMM algorithm for $\bu_i$ updating}
\label{alg:ui-update}
\end{algorithm}

\section{Experiments}


Now that we have presented our GRMF model in detail, we turn to its experimental validation by conducting experiments using real data sets and evaluate the performance of GRMF.
The biggest concern of our algorithm is to achieve the grouping effect and sparsity simultaneously, while keeping the reconstruction error at a low level, and to demonstrate the robustness to outliers and corruption noise. 
We compare GRMF with several state-of-the-art matrix factorization methods, which include 
Truncated SVD~\citep{hansen1987truncatedsvd},
RPCA~\citep{candes2011robust}, RNMF~\citep{wen2018survey},
RPMF~\citep{wang2012probabilistic} and  GoDec+~\citep{guo2017godec+}.
GRMF is our main method with $\ell_1$-loss, GMF-L2 is its variant with $\ell_2$-loss. 
We first study the behaviors of the 7 algorithms against different ranks and corruption ratios. We then apply all the algorithms to the face recovery problem. 
The desktop computer used to run our experiments has 2.7 GHz 64-bit Intel Core i5 processor (with two cores) and 8GB RAM.

\subsection{Performance under different corruption ratios and reduced ranks}

\begin{figure*}[tb]
\subfloat[\centering reconstruction error vs. corruption ratio ]{\includegraphics[width=0.45\textwidth]{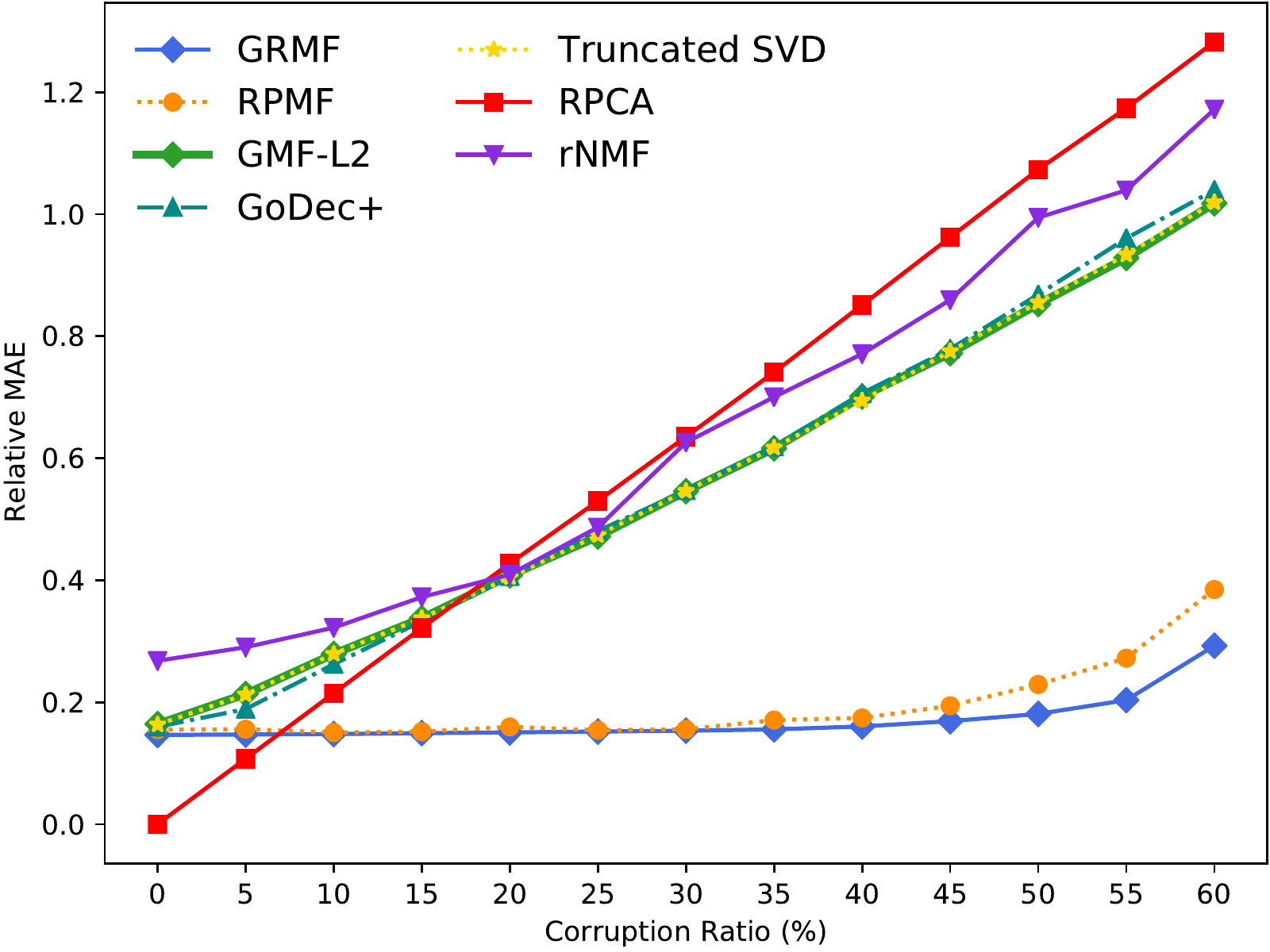} \label{fig:Error vs. Corruption Ratio} }%
\subfloat[\centering reconstruction error vs. reduced rank ]{\includegraphics[width=0.45\textwidth]{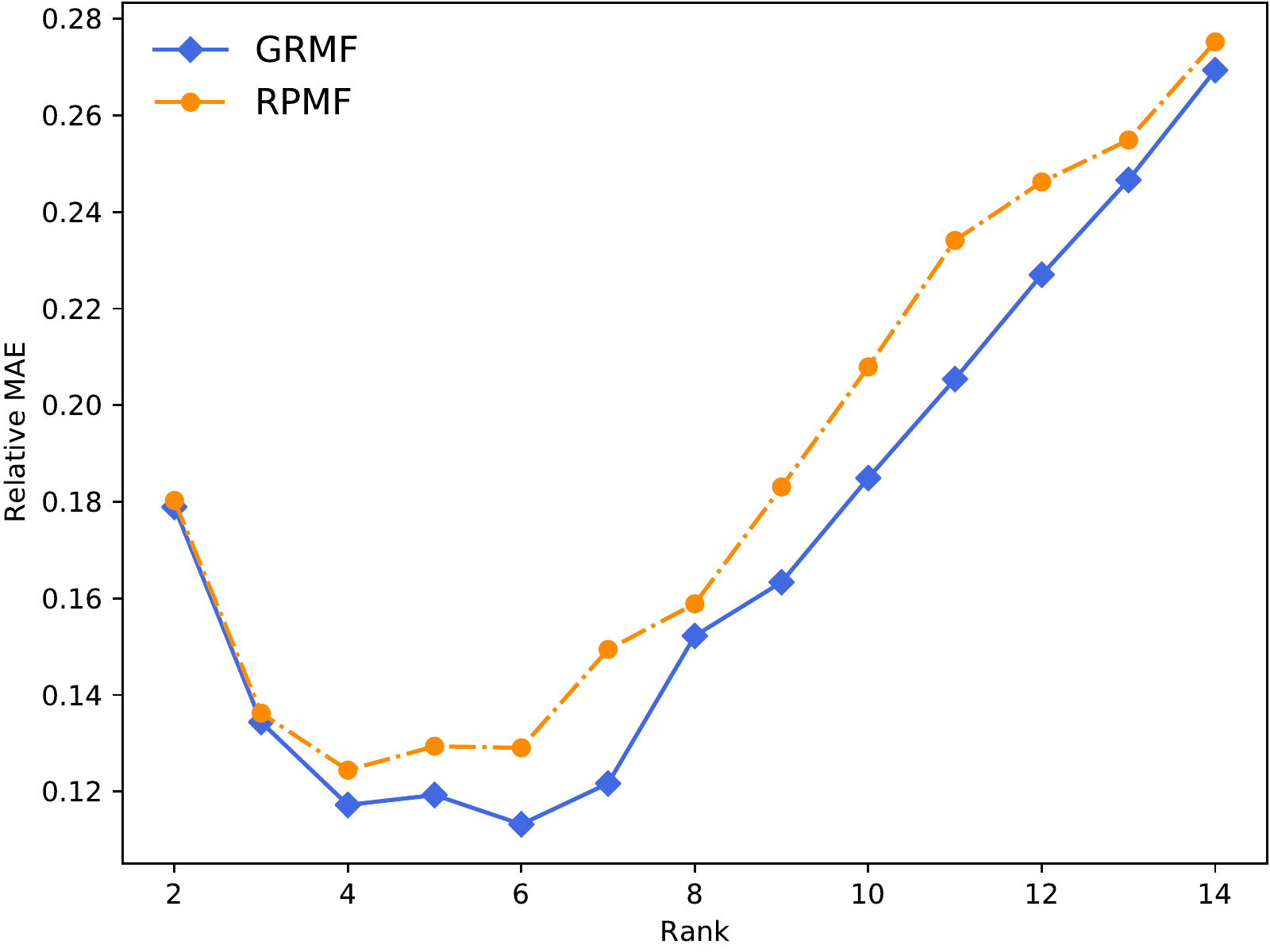} \label{fig:Error vs. Ranks} }%
\\
\subfloat[\centering Number of groups vs. reduced rank ]{\includegraphics[width=0.45\textwidth]{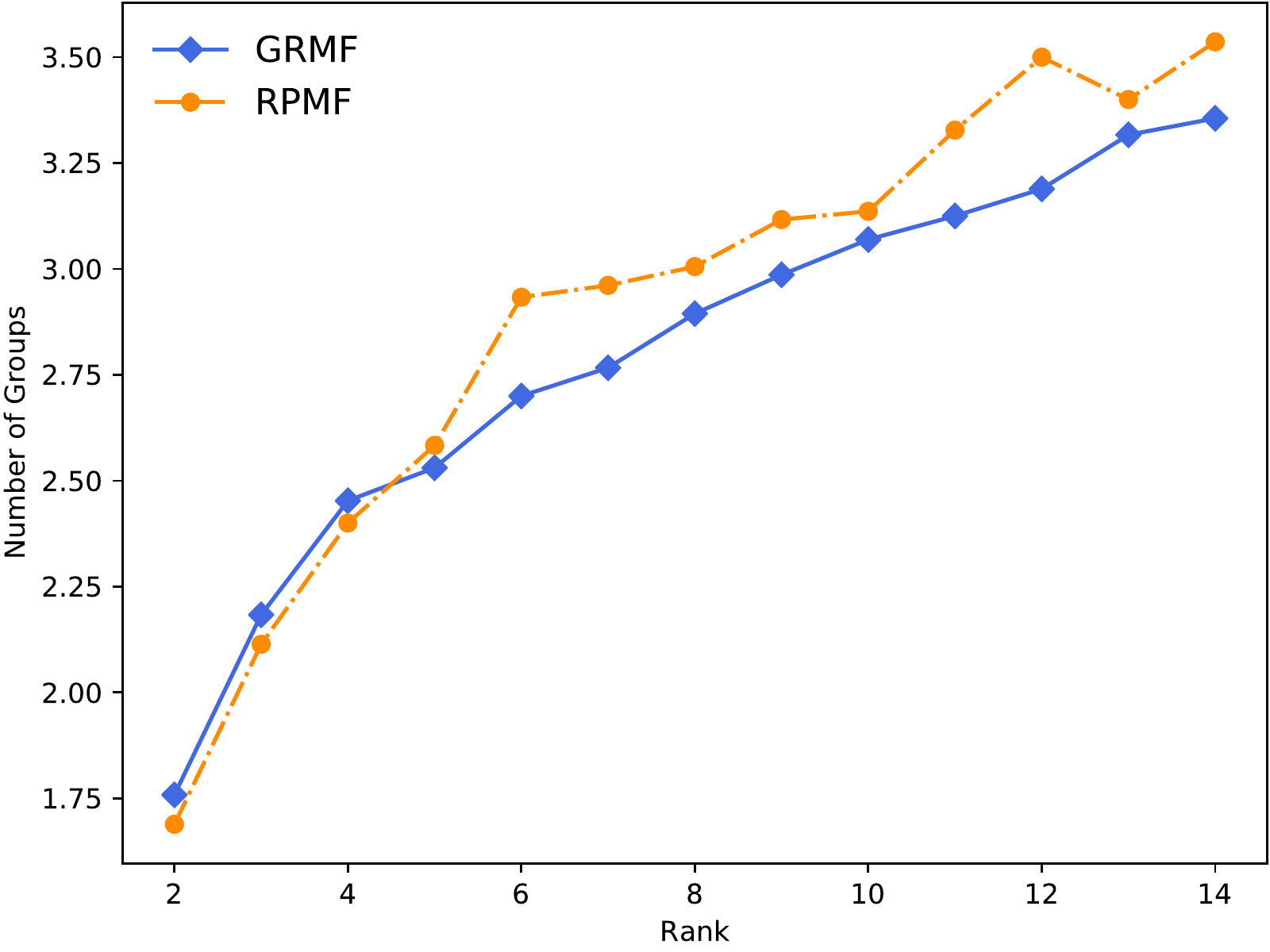} \label{fig:Number of groups of the main algorithms} }%
\quad
\subfloat[\centering Sparsity vs. reduced rank ]{\includegraphics[width=0.45\textwidth]{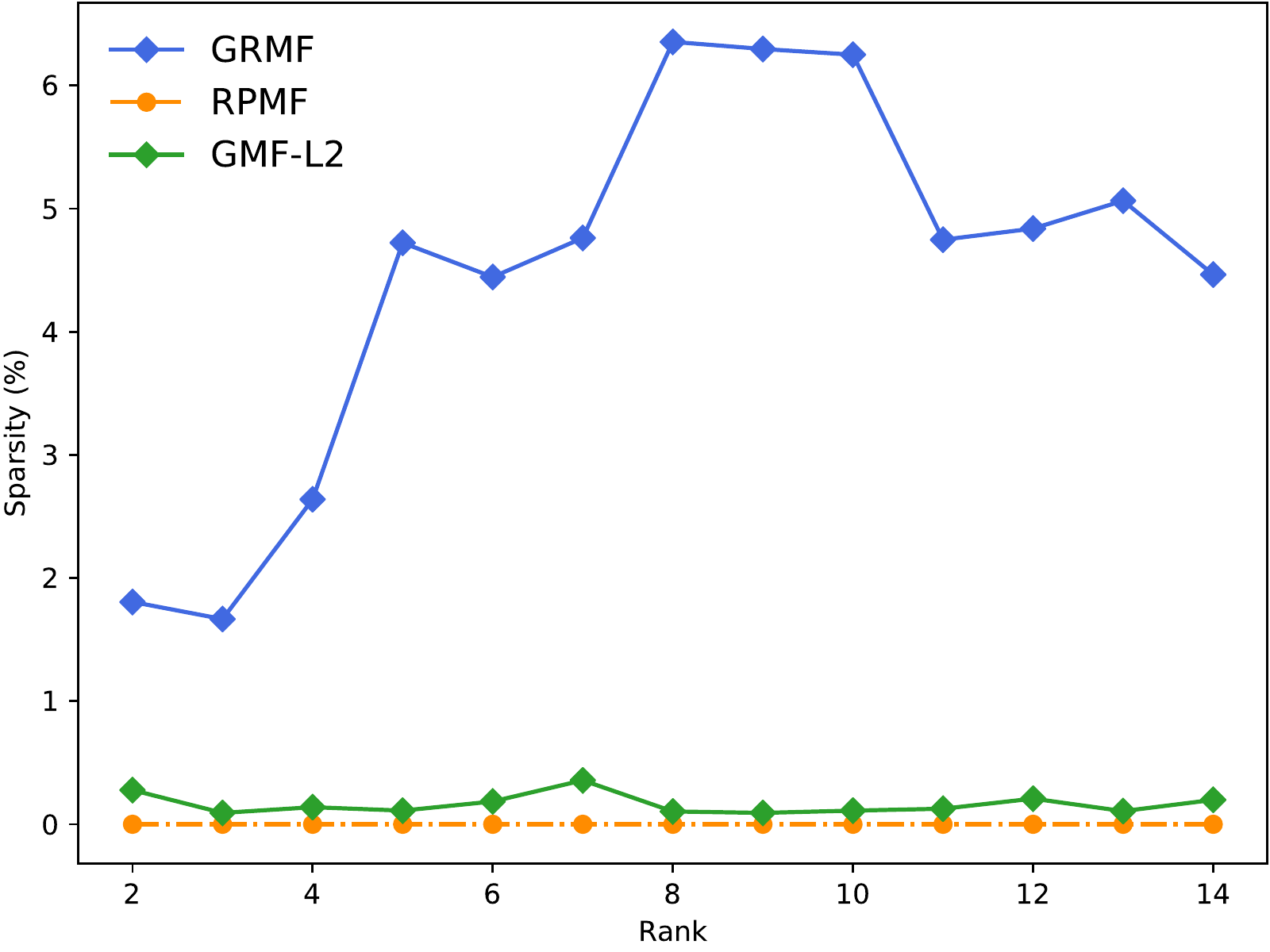} \label{fig:Sparsity of the main algorithms} }%
\caption{Performance under different corruption ratios and reduced ranks.}
\end{figure*}


The reduced rank and the corruption ratio can have a strong impact on the performance and the robustness of matrix factorization methods.
Thus the performance caused by these two factors should be thoroughly investigated in comparison of different matrix factorization methods. 
In this section, we show the behaviors of the 7 algorithms against different ranks and corruption ratios with a pilot experiment of one randomly chosen image from the Yale B dataset.

\textbf{Robustness vs. corruption ratio}\quad
In order to explore the robustness of the algorithms with the corruption ratio, we track the trajectory of the reconstruction errors with varying corruption ratios, while fixing the reduced rank.
The reduced rank of 4 is used to demonstrate the performance of different MF methods, as too large a value might cause overfitting.
The relative mean absolute error (Relative MAE) is used as a measure of reconstruction error, which is calculated by comparing the original image with the approximated image using the formula $\frac{\|\bY - \bU\bV^T\|_{\ell_1}}{\|\bY\|_{\ell_1}}$. 
Since the Relative MAE uses the $\ell_1$-norm loss, it can be viewed as the criterion to measure the robustness of different MF methods.
The type of corruption is salt and pepper noise, which is commonly used in computer vision applications. We add it to the original image by randomly masking part of the pixels in the image with the value of $0$ or $255$ under given corruption ratio.
For example, a $50\%$ corruption ratio means half of the pixels in a image is replaced by either $0$ or $255$ with equal probability.

Figure~\ref{fig:Error vs. Corruption Ratio} shows that both GRMF and RPMF can maintain the robustness with the lowest reconstruction errors, even when the corruption ratio gets very large.
Their errors increase slowly with the increase of the corruption ratio, and even when half of the pixels are corrupted with a corruption ratio of $50\%$ their errors will only increase slightly.
In particular, when the corruption ratio between the corrupted image and the original image is above $50\%$, GRMF and RPMF can still reduce the reconstruction error to around $0.2$.
It is also worth mentioning that RPCA can always recover the raw input images perfectly, but 
it does not have the ability to deal with the corrupted images with noise. 
This behavior results from the formulation of RPCA that it is not modeling the latent factors, and that it does not take rank as a parameter, which also explains that RPCA keeps almost all the information of the input image and its error increases linearly with the corruption ratio.

\textbf{Robustness vs. reduced rank}\quad
In order to explore the robustness of the algorithms with the reduced rank, we track the trajectory of the reconstruction errors with varying 
reduced ranks, while fixing the corruption ratio.
The corruption ratio of $50\%$ salt and pepper noise is used to demonstrate the performance of different MF methods, as most of the other MF methods fail to reconstruct the image except for GRMF and RPMF.

We run GRMF and RPMF with different ranks, since they are the only two algorithms that could well recover the original image under $50\%$ corruption ratio.
The results from Figure~\ref{fig:Error vs. Ranks} show that the reconstruction error remains at a low level when the reduced ranks are relatively small, and it first decreases then starts to increase as the reduced rank gets larger and larger, which shows a potential of overfitting.
Specifically, the reconstruction error of RPMF starts to increase when the reduced rank is bigger than $5$, while the error of GRMF starts to increase when the reduced rank is bigger than $7$, which shows that RPMF starts to overfit much earlier than GRMF.
Obviously, the grouping effect of GRMF helps denoising the factorization and makes it less sensitive to the choice of the reduced rank.
However, both GRMF and RPMF learn the noise and overfit the image as the rank gets too large. 
In addition, GRMF and RPMF have similar performance regarding the error, but GRMF has stronger grouping effect and sparsity, as illustrated in Figure~\ref{fig:Number of groups of the main algorithms} and Figure~\ref{fig:Sparsity of the main algorithms}. GRMF tends to have less number of groups and higher sparsity than RPMF.

\subsection{Image recovery}

\begin{table}
\centering 
\caption{Datasets information. We randomly choose some images of the same person/item, as the images for the same objective are very similar except in different angles/expressions/light conditions. The optimal rank for each dataset is reported on the fourth row. For the corrupted images, the rank is chosen to be one less than the optimal rank to avoid fitting the noise.\\}
\label{tab:datainfor}
\begin{tabular}{ccccc}
	\toprule
	Datasets & Yale B & COIL20 & ORL & JAFFE\\ 
	\hline
	Number of images & 210 & 206 & 200 & 213\\
	Size of images & 192$\times$168 & 128$\times$128 & 112$\times$92 & 180$\times$150\\
	Optimal rank & 5 & 5 & 4 & 5\\
	\bottomrule
\end{tabular}
\end{table}

Now we consider the application to image recovery in computer vision. We apply all the 7 algorithms on 4 datasets, Extended Yale B~\citep{GeBeKr01}, COIL20~\citep{Nene96columbiaobject}, ORL~\citep{Samaria94parameterisationof}, and JAFFE~\citep{lyonsmichael1998}. 
To avoid overfitting, we try several ranks and find an optimal rank in each situation. 
\autoref{tab:datainfor} reports the basic information about the datasets. 
We conduct the experiment on the original image and its $50\%$ corrupted version. 
The results on~\autoref{tab:main-result} show that our GRMF algorithm has a better recovery ability than the other algorithms, especially under severe corruption.
After running experiment over all the images in each dataset, we report the mean and the standard deviation of the relative MAE of the seven algorithms. All the algorithms perform well when there is no corruption, but only GRMF and PRMF remain at a low level with respect to the reconstruction error under $50\%$ corruption. GRMF has the lowest reconstruction error under all cases while exhibiting a low standard deviation. 
On average, each image costs less than 300s. 
And the parallelizable nature of our algorithm allows a multiprocessing or multithreding settings to accelerate the computation. More results are reported on {\color{blue}Appendix C}.

\begin{table*}[tbh]
\centering
\caption{Comparison of the reconstruction errors on the 4 datasets. The average of the relative mean absolute error (RMAE) is reported with the standard deviation in the parentheses. For the corrupted images, we randomly masked $50\%$ of the pixels in each image with salt and pepper noise. Note that we report RPCA only on the corrupted case. T-SVD stands for Truncated SVD.}

\label{tab:main-result}
\setlength\tabcolsep{2pt}  

\resizebox{\textwidth}{!}{  

\begin{tabular}{ccccccccc}
\toprule
\multirow{2}{*}{Methods} & \multicolumn{2}{c}{Yale B} & \multicolumn{2}{c}{COIL} & \multicolumn{2}{c}{ORL} & \multicolumn{2}{c}{JAFFE}\\

\cmidrule(r){2-3} \cmidrule(r){4-5} \cmidrule(r){6-7}  \cmidrule(r){8-9}

&  Origin  &  Corrupted
&  Origin  &  Corrupted
&  Origin  &  Corrupted
&  Origin  &  Corrupted \\

\midrule

GRMF &  
\textbf{0.093}$\pm$(0.022) & \textbf{0.143}$\pm$(0.041) & \textbf{0.123}$\pm$(0.066) & \textbf{0.245}$\pm$(0.135) & \textbf{0.105}$\pm$(0.020) & \textbf{0.204}$\pm$(0.036) & \textbf{0.121}$\pm$(0.013) & \textbf{0.165}$\pm$(0.018) \\

PRMF & 
0.095$\pm$(0.023) & 0.154$\pm$(0.047) & 0.127$\pm$(0.070) & 0.273$\pm$(0.142) & 0.107$\pm$(0.020) & 0.210$\pm$(0.037) & 0.125$\pm$(0.014) & 0.182$\pm$(0.025)  \\

GMF-L2 & 
0.103$\pm$(0.026) & 0.555$\pm$(0.367) & 0.143$\pm$(0.075) & 0.821$\pm$(0.467) & 0.114$\pm$(0.021) & 0.320$\pm$(0.052) & 0.133$\pm$(0.013) & 0.398$\pm$(0.055)  \\

GoDec+ & 
0.101$\pm$(0.025) & 0.565$\pm$(0.370) & 0.139$\pm$(0.074) & 0.830$\pm$(0.470) & 0.113$\pm$(0.022) & 0.325$\pm$(0.053) & 0.132$\pm$(0.013) & 0.412$\pm$(0.058) \\

T-SVD 
& 0.103$\pm$(0.026) & 0.560$\pm$(0.368) & 0.143$\pm$(0.075) & 0.824$\pm$(0.469) & 0.114$\pm$(0.021) & 0.324$\pm$(0.052) & 0.133$\pm$(0.013) & 0.401$\pm$(0.056)  \\

RPCA & 
- & 0.835$\pm$(0.380) & - & 0.963$\pm$(0.446) & - & 0.577$\pm$(0.073) & - & 0.608$\pm$(0.078)  \\

RNMF & 
0.149$\pm$(0.053) & 0.626$\pm$(0.393) & 0.285$\pm$(0.280) & 0.722$\pm$(0.214) & 0.130$\pm$(0.025) & 0.360$\pm$(0.055) & 0.207$\pm$(0.022) & 0.456$\pm$(0.064)  \\

\bottomrule
\end{tabular}
}
\end{table*}

\section{Conclusion}

In this work, we studied the problem of matrix factorization incorporating sparsity and grouping effect. We proposed a novel method, namely \textit{Robust Matrix Factorization with Grouping effect} (GRMF), which intends to lower the reconstruction errors while promoting intepretability of the solution through automatically determining the number of latent groups in the solution. To the best of our knowledge, it is the first paper to introduce the automatic learning of grouping effect without prior information into MF. Specifically, GRMF incorporates two novel non-convex regularizers that control both sparsity and grouping effect in the objective function, and a novel optimization framework is proposed to obtain the solution. Moreover, GRMF employs an alternative minimization procedure to decompose the problem into a number of coupled non-convex subproblems, where each subproblem optimizes a row or a column in the solution of MF through difference-of-convex (DC) programming and the alternating direction method of multipliers (ADMM). We have conducted extensive experiments to evaluate GRMF using (1) Extended Yale B dataset under different rank choices and corruption ratios and (2) image reconstruction tasks on 4 image datasets (COIL-20, ORL, Extended Yale B, and JAFFE) under $50\%$ corruption ratio. Compared with 6 baseline algorithms, GRMF has achieved the best reconstruction accuracy in both tasks, while demonstrating the performance advances from the use of grouping effects and sparsity, especially under severe data corruption.

\bibliographystyle{plainnat}  
\bibliography{ref.bib}

\begin{thebibliography}{49}
\providecommand{\natexlab}[1]{#1}
\providecommand{\url}[1]{\texttt{#1}}
\expandafter\ifx\csname urlstyle\endcsname\relax
  \providecommand{\doi}[1]{doi: #1}\else
  \providecommand{\doi}{doi: \begingroup \urlstyle{rm}\Url}\fi

\bibitem[Abdolali and Gillis(2021)]{abdolali2021simplex}
Maryam Abdolali and Nicolas Gillis.
\newblock Simplex-structured matrix factorization: Sparsity-based
  identifiability and provably correct algorithms.
\newblock \emph{SIAM Journal on Mathematics of Data Science}, 3\penalty0
  (2):\penalty0 593--623, 2021.

\bibitem[Cand{\`e}s et~al.(2011)Cand{\`e}s, Li, Ma, and
  Wright]{candes2011robust}
Emmanuel~J Cand{\`e}s, Xiaodong Li, Yi~Ma, and John Wright.
\newblock Robust principal component analysis?
\newblock \emph{Journal of the ACM (JACM)}, 58\penalty0 (3):\penalty0 1--37,
  2011.

\bibitem[Chi et~al.(2019)Chi, Lu, and Chen]{chi2019nonconvex}
Yuejie Chi, Yue~M Lu, and Yuxin Chen.
\newblock Nonconvex optimization meets low-rank matrix factorization: An
  overview.
\newblock \emph{IEEE Transactions on Signal Processing}, 67\penalty0
  (20):\penalty0 5239--5269, 2019.

\bibitem[Choi(2008)]{choi2008algorithms}
Seungjin Choi.
\newblock Algorithms for orthogonal nonnegative matrix factorization.
\newblock In \emph{2008 ieee international joint conference on neural networks
  (ieee world congress on computational intelligence)}, pages 1828--1832. IEEE,
  2008.

\bibitem[Csisz{\'{a}}r and Tusn{\'{a}}dy(1984)]{Csiszar}
I.~Csisz{\'{a}}r and G.~Tusn{\'{a}}dy.
\newblock Information geometry and alternating minimization procedures.
\newblock \emph{Statistics \& Decisions}, \penalty0 (1):\penalty0 205--237,
  1984.

\bibitem[Cui et~al.(2019)Cui, Liu, Gao, Zheng, and Wang]{cui2019rcmf}
Zhen Cui, Jin-Xing Liu, Ying-Lian Gao, Chun-Hou Zheng, and Juan Wang.
\newblock Rcmf: a robust collaborative matrix factorization method to predict
  mirna-disease associations.
\newblock \emph{BMC bioinformatics}, 20\penalty0 (25):\penalty0 1--10, 2019.

\bibitem[Dai et~al.(2020)Dai, Su, Zhang, Xue, and Li]{dai2020robust}
Xiangguang Dai, Xiaojie Su, Wei Zhang, Fangzheng Xue, and Huaqing Li.
\newblock Robust manhattan non-negative matrix factorization for image recovery
  and representation.
\newblock \emph{Information Sciences}, 527:\penalty0 70--87, 2020.

\bibitem[Gao et~al.(2022)Gao, Guo, Zhu, Kan, and Zhang]{gao2022human}
Hongbo Gao, Fang Guo, Juping Zhu, Zhen Kan, and Xinyu Zhang.
\newblock Human motion segmentation based on structure constraint matrix
  factorization.
\newblock \emph{Inform. Sci}, 2022\penalty0 (65):\penalty0 119103, 2022.

\bibitem[Gaujoux and Seoighe(2010)]{gaujoux2010flexible}
Renaud Gaujoux and Cathal Seoighe.
\newblock A flexible r package for nonnegative matrix factorization.
\newblock \emph{BMC bioinformatics}, 11\penalty0 (1):\penalty0 1--9, 2010.

\bibitem[Georghiades et~al.(2001)Georghiades, Belhumeur, and
  Kriegman]{GeBeKr01}
A.S. Georghiades, P.N. Belhumeur, and D.J. Kriegman.
\newblock From few to many: Illumination cone models for face recognition under
  variable lighting and pose.
\newblock \emph{IEEE Trans. Pattern Anal. Mach. Intelligence}, 23\penalty0
  (6):\penalty0 643--660, 2001.

\bibitem[Guo et~al.(2017)Guo, Liu, Xu, Xu, and Tao]{guo2017godec+}
Kailing Guo, Liu Liu, Xiangmin Xu, Dong Xu, and Dacheng Tao.
\newblock Godec+: Fast and robust low-rank matrix decomposition based on
  maximum correntropy.
\newblock \emph{IEEE transactions on neural networks and learning systems},
  29\penalty0 (6):\penalty0 2323--2336, 2017.

\bibitem[Gurini et~al.(2018)Gurini, Gasparetti, Micarelli, and
  Sansonetti]{gurini2018temporal}
Davide~Feltoni Gurini, Fabio Gasparetti, Alessandro Micarelli, and Giuseppe
  Sansonetti.
\newblock Temporal people-to-people recommendation on social networks with
  sentiment-based matrix factorization.
\newblock \emph{Future Generation Computer Systems}, 78:\penalty0 430--439,
  2018.

\bibitem[Haeffele et~al.(2014)Haeffele, Young, and
  Vidal]{haeffele2014structured}
Benjamin Haeffele, Eric Young, and Rene Vidal.
\newblock Structured low-rank matrix factorization: Optimality, algorithm, and
  applications to image processing.
\newblock In \emph{International conference on machine learning}, pages
  2007--2015. PMLR, 2014.

\bibitem[Haeffele and Vidal(2019)]{haeffele2019structured}
Benjamin~D Haeffele and Ren{\'e} Vidal.
\newblock Structured low-rank matrix factorization: Global optimality,
  algorithms, and applications.
\newblock \emph{IEEE transactions on pattern analysis and machine
  intelligence}, 42\penalty0 (6):\penalty0 1468--1482, 2019.

\bibitem[Hansen(1987)]{hansen1987truncatedsvd}
Per~Christian Hansen.
\newblock The truncatedsvd as a method for regularization.
\newblock \emph{BIT Numerical Mathematics}, 27\penalty0 (4):\penalty0 534--553,
  1987.

\bibitem[Hardt(2014)]{hardt2014understanding}
Moritz Hardt.
\newblock Understanding alternating minimization for matrix completion.
\newblock In \emph{2014 IEEE 55th Annual Symposium on Foundations of Computer
  Science}, pages 651--660. IEEE, 2014.

\bibitem[Hoyer(2004)]{hoyer2004non}
Patrik~O Hoyer.
\newblock Non-negative matrix factorization with sparseness constraints.
\newblock \emph{Journal of machine learning research}, 5\penalty0 (9), 2004.

\bibitem[Jakomin et~al.(2020)Jakomin, Bosni{\'c}, and
  Curk]{jakomin2020simultaneous}
Martin Jakomin, Zoran Bosni{\'c}, and Toma{\v{z}} Curk.
\newblock Simultaneous incremental matrix factorization for streaming
  recommender systems.
\newblock \emph{Expert Systems with Applications}, 160:\penalty0 113685, 2020.

\bibitem[Jamali et~al.(2020)Jamali, Kusalik, and Wu]{jamali2020mdipa}
Ali~Akbar Jamali, Anthony Kusalik, and Fang-Xiang Wu.
\newblock Mdipa: a microrna--drug interaction prediction approach based on
  non-negative matrix factorization.
\newblock \emph{Bioinformatics}, 36\penalty0 (20):\penalty0 5061--5067, 2020.

\bibitem[Kim et~al.(2012)Kim, Monteiro, and Park]{kim2012group}
Jingu Kim, Renato~DC Monteiro, and Haesun Park.
\newblock Group sparsity in nonnegative matrix factorization.
\newblock In \emph{Proceedings of the 2012 SIAM International Conference on
  Data Mining}, pages 851--862. SIAM, 2012.

\bibitem[Klema and Laub(1980)]{klema1980singular}
Virginia Klema and Alan Laub.
\newblock The singular value decomposition: Its computation and some
  applications.
\newblock \emph{IEEE Transactions on automatic control}, 25\penalty0
  (2):\penalty0 164--176, 1980.

\bibitem[Koren et~al.(2009)Koren, Bell, and Volinsky]{koren2009matrix}
Yehuda Koren, Robert Bell, and Chris Volinsky.
\newblock Matrix factorization techniques for recommender systems.
\newblock \emph{Computer}, 42\penalty0 (8):\penalty0 30--37, 2009.

\bibitem[Li et~al.(2013)Li, Xu, Cao, Fan, and Niu]{li2013cgmf}
Fangfang Li, Guandong Xu, Longbing Cao, Xiaozhong Fan, and Zhendong Niu.
\newblock Cgmf: Coupled group-based matrix factorization for recommender
  system.
\newblock In \emph{International Conference on Web Information Systems
  Engineering}, pages 189--198. Springer, 2013.

\bibitem[Li et~al.(2019)Li, Zhang, and Du]{li2019robust}
Ruyue Li, Lefei Zhang, and Bo~Du.
\newblock A robust dimensionality reduction and matrix factorization framework
  for data clustering.
\newblock \emph{Pattern Recognition Letters}, 128:\penalty0 440--446, 2019.

\bibitem[Lin et~al.(2017)Lin, Xu, and Zha]{lin2017robust}
Zhouchen Lin, Chen Xu, and Hongbin Zha.
\newblock Robust matrix factorization by majorization minimization.
\newblock \emph{IEEE transactions on pattern analysis and machine
  intelligence}, 40\penalty0 (1):\penalty0 208--220, 2017.

\bibitem[Lyons et~al.(1998)Lyons, Kamachi, and Gyoba]{lyonsmichael1998}
Michael Lyons, Miyuki Kamachi, and Jiro Gyoba.
\newblock {The Japanese Female Facial Expression (JAFFE) Dataset}, April 1998.
\newblock URL \url{https://doi.org/10.5281/zenodo.3451524}.
\newblock {The images are provided at no cost for non- commercial scientific
  research only. If you agree to the conditions listed below, you may request
  access to download.}

\bibitem[Ma et~al.(2008)Ma, Yang, Lyu, and King]{ma2008sorec}
Hao Ma, Haixuan Yang, Michael~R Lyu, and Irwin King.
\newblock Sorec: social recommendation using probabilistic matrix
  factorization.
\newblock In \emph{Proceedings of the 17th ACM conference on Information and
  knowledge management}, pages 931--940, 2008.

\bibitem[Mnih and Salakhutdinov(2007)]{mnih2007probabilistic}
Andriy Mnih and Russ~R Salakhutdinov.
\newblock Probabilistic matrix factorization.
\newblock \emph{Advances in neural information processing systems},
  20:\penalty0 1257--1264, 2007.

\bibitem[Na et~al.(2019)Na, Kang, Jung, and Kang]{na2019nonconvex}
Hanwool Na, Myeongmin Kang, Miyoun Jung, and Myungjoo Kang.
\newblock Nonconvex tgv regularization model for multiplicative noise removal
  with spatially varying parameters.
\newblock \emph{Inverse Problems \& Imaging}, 13\penalty0 (1):\penalty0 117,
  2019.

\bibitem[Nene et~al.(1996)Nene, Nayar, and Murase]{Nene96columbiaobject}
Sameer~A. Nene, Shree~K. Nayar, and Hiroshi Murase.
\newblock Columbia object image library (coil-20.
\newblock Technical report, 1996.

\bibitem[Ortega et~al.(2016)Ortega, Hernando, Bobadilla, and
  Kang]{ortega2016recommending}
Fernando Ortega, Antonio Hernando, Jesus Bobadilla, and Jeon~Hyung Kang.
\newblock Recommending items to group of users using matrix factorization based
  collaborative filtering.
\newblock \emph{Information Sciences}, 345:\penalty0 313--324, 2016.

\bibitem[Parvin et~al.(2019)Parvin, Moradi, Esmaeili, and
  Qader]{parvin2019scalable}
Hashem Parvin, Parham Moradi, Shahrokh Esmaeili, and Nooruldeen~Nasih Qader.
\newblock A scalable and robust trust-based nonnegative matrix factorization
  recommender using the alternating direction method.
\newblock \emph{Knowledge-Based Systems}, 166:\penalty0 92--107, 2019.

\bibitem[Pascual-Montano et~al.(2006)Pascual-Montano, Carmona-Saez, Chagoyen,
  Tirado, Carazo, and Pascual-Marqui]{pascual2006bionmf}
Alberto Pascual-Montano, Pedro Carmona-Saez, Monica Chagoyen, Francisco Tirado,
  Jose~M Carazo, and Roberto~D Pascual-Marqui.
\newblock bionmf: a versatile tool for non-negative matrix factorization in
  biology.
\newblock \emph{BMC bioinformatics}, 7\penalty0 (1):\penalty0 1--9, 2006.

\bibitem[Rahimpour et~al.(2017)Rahimpour, Qi, Fugate, and
  Kuruganti]{rahimpour2017non}
Alireza Rahimpour, Hairong Qi, David Fugate, and Teja Kuruganti.
\newblock Non-intrusive energy disaggregation using non-negative matrix
  factorization with sum-to-k constraint.
\newblock \emph{IEEE Transactions on Power Systems}, 32\penalty0 (6):\penalty0
  4430--4441, 2017.

\bibitem[Samaria et~al.(1994)Samaria, *t, Harter, and
  Site]{Samaria94parameterisationof}
F.~S. Samaria, F.~S.~Samaria *t, A.C. Harter, and Old~Addenbrooke's Site.
\newblock Parameterisation of a stochastic model for human face identification,
  1994.

\bibitem[Shen et~al.(2012)Shen, Pan, and Zhu]{shen2012likelihood}
Xiaotong Shen, Wei Pan, and Yunzhang Zhu.
\newblock Likelihood-based selection and sharp parameter estimation.
\newblock \emph{Journal of the American Statistical Association}, 107\penalty0
  (497):\penalty0 223--232, 2012.

\bibitem[Srebro et~al.(2005)Srebro, Rennie, and Jaakkola]{srebro2005maximum}
Nathan Srebro, Jason Rennie, and Tommi~S Jaakkola.
\newblock Maximum-margin matrix factorization.
\newblock In \emph{Advances in neural information processing systems}, pages
  1329--1336, 2005.

\bibitem[Wang and Yeung(2013)]{wang2013bayesian}
Naiyan Wang and Dit-Yan Yeung.
\newblock Bayesian robust matrix factorization for image and video processing.
\newblock In \emph{Proceedings of the IEEE International Conference on Computer
  Vision}, pages 1785--1792, 2013.

\bibitem[Wang et~al.(2012)Wang, Yao, Wang, and Yeung]{wang2012probabilistic}
Naiyan Wang, Tiansheng Yao, Jingdong Wang, and Dit-Yan Yeung.
\newblock A probabilistic approach to robust matrix factorization.
\newblock In \emph{European Conference on Computer Vision}, pages 126--139.
  Springer, 2012.

\bibitem[Wang et~al.(2020)Wang, He, Jiang, and Li]{wang2020robust}
Qi~Wang, Xiang He, Xu~Jiang, and Xuelong Li.
\newblock Robust bi-stochastic graph regularized matrix factorization for data
  clustering.
\newblock \emph{IEEE Transactions on Pattern Analysis and Machine
  Intelligence}, 2020.

\bibitem[Wen et~al.(2018)Wen, Chu, Liu, and Qiu]{wen2018survey}
Fei Wen, Lei Chu, Peilin Liu, and Robert~C Qiu.
\newblock A survey on nonconvex regularization-based sparse and low-rank
  recovery in signal processing, statistics, and machine learning.
\newblock \emph{IEEE Access}, 6:\penalty0 69883--69906, 2018.

\bibitem[Wold et~al.(1987)Wold, Esbensen, and Geladi]{wold1987principal}
Svante Wold, Kim Esbensen, and Paul Geladi.
\newblock Principal component analysis.
\newblock \emph{Chemometrics and intelligent laboratory systems}, 2\penalty0
  (1-3):\penalty0 37--52, 1987.

\bibitem[Xu et~al.(2020)Xu, Zhang, and Zhang]{xu2020bayesian}
Shuang Xu, Chunxia Zhang, and Jiangshe Zhang.
\newblock Bayesian deep matrix factorization network for multiple images
  denoising.
\newblock \emph{Neural Networks}, 123:\penalty0 420--428, 2020.

\bibitem[Xue et~al.(2017)Xue, Dai, Zhang, Huang, and Chen]{xue2017deep}
Hong-Jian Xue, Xinyu Dai, Jianbing Zhang, Shujian Huang, and Jiajun Chen.
\newblock Deep matrix factorization models for recommender systems.
\newblock In \emph{IJCAI}, volume~17, pages 3203--3209. Melbourne, Australia,
  2017.

\bibitem[Yang et~al.(2012)Yang, Yuan, Lai, Shen, Wonka, and
  Ye]{yang2012feature}
Sen Yang, Lei Yuan, Ying-Cheng Lai, Xiaotong Shen, Peter Wonka, and Jieping Ye.
\newblock Feature grouping and selection over an undirected graph.
\newblock In \emph{Proceedings of the 18th ACM SIGKDD international conference
  on Knowledge discovery and data mining}, pages 922--930, 2012.

\bibitem[Yao and Kwok(2018)]{yao2017scalable}
Quanming Yao and James Kwok.
\newblock Scalable robust matrix factorization with nonconvex loss.
\newblock \emph{arXiv preprint arXiv:1710.07205}, 31, 2018.
\newblock URL
  \url{https://proceedings.neurips.cc/paper/2018/file/2c3ddf4bf13852db711dd1901fb517fa-Paper.pdf}.

\bibitem[Yuen et~al.(2012)Yuen, King, and Leung]{yuen2012taskrec}
Man-Ching Yuen, Irwin King, and Kwong-Sak Leung.
\newblock Taskrec: probabilistic matrix factorization in task recommendation in
  crowdsourcing systems.
\newblock In \emph{International Conference on Neural Information Processing},
  pages 516--525. Springer, 2012.

\bibitem[Zhang et~al.(2011)Zhang, Chen, Zheng, and He]{zhang2011robust}
Lijun Zhang, Zhengguang Chen, Miao Zheng, and Xiaofei He.
\newblock Robust non-negative matrix factorization.
\newblock \emph{Frontiers of Electrical and Electronic Engineering in China},
  6\penalty0 (2):\penalty0 192--200, 2011.

\bibitem[Zhang et~al.(2020)Zhang, Yun, Dai, Cui, and Shang]{zhang2020graphs}
Yupei Zhang, Yue Yun, Huan Dai, Jiaqi Cui, and Xuequn Shang.
\newblock Graphs regularized robust matrix factorization and its application on
  student grade prediction.
\newblock \emph{Applied Sciences}, 10\penalty0 (5):\penalty0 1755, 2020.

\end{thebibliography}

\newpage
\appendix
\appendixpage
\addappheadtotoc

\subsection*{Illustration of the grouping effect of GRMF}

\begin{figure}[H]
\[
    \begin{tikzpicture}
        \matrix [matrix of math nodes,left delimiter=(,right delimiter=)] (m)
        {
            100  &  110  &  90  & \dots &200  &  190  &  210   \\
            100  &  110  &  90  & \dots &200  &  190  &  210   \\
            \vdots & \vdots & \vdots & ~ & \vdots & \vdots & \vdots\\
            0.01  &  100  &  120  & \dots & 2.5    &  0.21  &  90   \\
        };
    \end{tikzpicture}
    \xLongrightarrow{\text{Grouping}}
    \begin{tikzpicture}
        \matrix [matrix of math nodes,left delimiter=(,right delimiter=)] (m)
        {
            |[draw=blue,fill=blue!20,outer sep=0,inner sep=2pt]|101 
            &|[draw=blue,fill=blue!20,outer sep=1,inner sep=2pt]|99.5
            &|[draw=blue,fill=blue!20,outer sep=0,inner sep=2pt]|100 & \dots &|[draw=green,fill=green!20,outer sep=0,inner sep=2pt]|200 
            &|[draw=green,fill=green!20,outer sep=0,inner sep=2pt]|199.5 
            &|[draw=green,fill=green!20,outer sep=0,inner sep=2pt]|201 \\
            |[draw=blue,fill=blue!20,outer sep=0,inner sep=2pt]|100 
            &|[draw=blue,fill=blue!20,outer sep=1,inner sep=2pt]|100.5 
            &|[draw=blue,fill=blue!20,outer sep=0,inner sep=2pt]|99.5 & \dots &|[draw=green,fill=green!20,outer sep=0,inner sep=2pt]|201 
            &|[draw=green,fill=green!20,outer sep=0,inner sep=2pt]|200 
            &|[draw=green,fill=green!20,outer sep=0,inner sep=2pt]|199 \\
            \vdots & \vdots & \vdots & ~ & \vdots & \vdots & \vdots\\
            |[draw=orange,fill=orange!20,outer sep=0,inner sep=2pt]|0.01  
            &|[draw=cyan,fill=cyan!20,outer sep=0,inner sep=2pt]|100 
            &|[draw=cyan,fill=cyan!20,outer sep=0,inner sep=2pt]|105 & \dots &|[draw=orange,fill=orange!20,outer sep=0,inner sep=2pt]|0.91 
            &|[draw=orange,fill=orange!20,outer sep=0,inner sep=2pt]|0.01 
            &|[draw=cyan,fill=cyan!20,outer sep=0,inner sep=2pt]|97.5 \\
        };
    \end{tikzpicture}
\]
\caption{Illustration of grouping effect}
\label{fig:grouping}
\end{figure}

We take the grouping effect in a matrix as an example. As shown in {\color{blue}Figure 2}, the grouping effect along each row of the matrix can introduce clustering between similar items, where elements close to each other are clustered in the same group which is highlighted with the same color. 
In addition, the elements clustered in the same group do not need to be adjacent, 
In the decomposed matrix obtained from GRMF, this grouping effect along the hidden factors is expected in both $\bU$ and $\bV$.

\section{Details of the algorithm of GRMF}
A Python implementation of GRMF can be found on github\footnote{https://github.com/GroupingEffects/GRMF}.

\subsection{Difference-of-convex algorithm (DC)}
Note that the optimization problem of $L(\bU|\bV)$ in~\eqref{eq:LU-fixV} can be decomposed into $d$ independent subproblems. The same procedure applies to the minimization of $L(\bV|\bU)$ in~\eqref{eq:LV-fixU}. Thus, the problem of GRMF is a combination of $n+d$ optimization subproblems, each with respect to a vector in $\mathbb{R}^r$. For each alternative minimization, we need to solve the following optimization problem: 
\begin{align*}
  \argmin_{\bx} \Bigg\{ S(\bx) 
    = &\sum_{i=1}^n [(b_i - \ba_i^T\bx)^2 + \epsilon]^{\frac{1}{2}} 
    + \lambda_1 \sum_{l=1}^r \min  \left(\frac{|x_l|}{\tau_1}, 1 \right)\\
    &+ \lambda_2 \sum_{l<l':(l,l')\in \mathcal{E}}^r \min \left(\frac{|x_l-x_{l'}|}{\tau_2}, 1 \right)
    + \lambda_3 \sum_{l=1}^{r} x_l^2 \Bigg\} \ .
\end{align*}

By applying that $\min (a,b) = a-(a-b)_{+}$, $S(\bx)$ can be decomposed into a difference of two convex functions as follows, 
\begin{align*}
  &S_1(\bx) = \sum_{i=1}^n [(b_i - \ba_i^T\bx)^2 + \epsilon]^{\frac{1}{2}} 
    + \lambda_1 \sum_{l=1}^r \frac{|x_l|}{\tau_1} 
    + \lambda_2 \sum_{l<l':(l,l')\in \mathcal{E}}^r \frac{|x_l-x_{l'}|}{\tau_2} 
    + \lambda_3 \sum_{l=1}^r x_l^2 , \\
  &S_2(\bx) = \lambda_1 \sum_{l=1}^r \left(\frac{|x_l|}{\tau_1} - 1\right)_{+} 
    + \lambda_2 \sum_{l<l':(l,l')\in \mathcal{E}}^r \left(\frac{|x_l-x_{l'}|}{\tau_2}-1\right)_{+} \ .
\end{align*}

\paragraph{Approximation of the trailing convex function}
We then construct a sequence of approximations of $S_2(\bx)$ iteratively. At the $m$-th iteration, we approximate $S_{2}(\bx)$ by its affine minorizaion, by linearizing the trailing convex function $S_{2}(\bx)$.

By linearizing the trailing convex function $S_2(\bx)$ in the difference convex decomposition, we replace $S_2(\bx)$ at the $m$-th iteration with its affine minorization at $(m-1)$-th iteration. We linearize the trailing convex function $S_2(\bmeta) = S_2(\bmeta^{*}) + \langle\bmeta - \bmeta^{*}, \partial S_2(\bmeta^{*}) \rangle$ at a neighborhood of $\bmeta^{*} \in \mathbb{R}^{(r^2+r)/2}$, where $\partial S_2(\bmeta)$ is the first derivative of $ S_2(\bmeta)$ with respect to $\bmeta$, and $\langle\cdot,\cdot\rangle$ is the inner product, where 
\begin{equation*}
  \bmeta = \left(|x_1|,|x_2|,\cdots,|x_r|,|x_{12}|,\cdots,|x_{1r}|,\cdots,|x_{(r-1)r}| \right)^T .
\end{equation*}
Then we get
\begin{equation*}
  S(\bmeta) = S(\bmeta^{*}) 
  + \langle\bmeta - \bmeta^{*}, \nabla S(\bmeta)\vert_{\bmeta = \bmeta^{*}}\rangle .
\end{equation*}

Thus, at the $m$-th iteration, we replace $S_2(\bx)$ with the $m$-th approximation by 
\begin{equation*}
  S_2^{(m)}(\bx) 
    = \Tilde{S}_2^{(m)}(\bmeta) 
    = \Tilde{S}_2(\widehat{\bmeta}^{(m-1)}) 
      + \langle\bmeta - \widehat{\bmeta}^{(m-1)}, \partial\Tilde{S}_2(\widehat{\bmeta}^{(m-1)}) \rangle .
\end{equation*}
Specifically,

\begin{equation}\label{eq:S_2 linearized}
\begin{split}
  S_2^{(m)}(\bx) 
    = & S_2(\widehat{\bx}^{(m-1)}) 
      + \langle\bx - \widehat{\bx}^{(m-1)}, 
        \left.\partial S_2(\bx)\right\vert_{\bx = \widehat{\bx}^{(m-1)}}\rangle\\
    = &S_2(\widehat{\bx}^{(m-1)}) 
      + \frac{\lambda_1}{\tau_1} \sum_{l=1}^r 
      I_{\left\{|\hat{x}_l^{(m-1)}|\geq \tau_1 \right\}}
        \cdot \left(|x_l| - |\hat{x}_l^{(m-1)}| \right)\\
      &+ \frac{\lambda_2}{\tau_2} \sum_{l<l': (l,l')\in \mathcal{E}} 
      I_{\left\{ |\hat{x}_{l}^{(m-1)} - \hat{x}_{l'}^{(m-1)}|\geq \tau_2 \right\}}
        \cdot \left(|x_l - x_{l'}| - |\hat{x}_{l}^{(m-1)} - \hat{x}_{l'}^{(m-1)}| \right) \ .
\end{split}
\end{equation}

Finally, a sequence of approximations of $S(\bx)$ is constructed iteratively. For the $m$-th approximation, an upper convex approximating function to $S(\bx)$ can be obtained by $S^{(m)}(\bx) = S_1(\bx) - S_2^{(m)}(\bx)$, which formulates the following subproblem.
\begin{equation} \label{eq:sub-problem}
  \min_{\bx} \sum_{i=1}^n [(b_i - \ba_i^T\bx)^2 + \epsilon]^{\frac{1}{2}} 
    + \frac{\lambda_1}{\tau_1} \sum_{l \in \mathcal{F}^{(m-1)}} |x_l| 
    + \frac{\lambda_2}{\tau_2} \sum_{l<l': (l,l')\in \mathcal{E}^{(m-1)}} |x_l - x_{l'}| 
    + \lambda_3 \sum_{l =1}^r x_l^2 \ .
\end{equation}
where 
\begin{equation}\label{eq:EF-update-appendix}
\begin{split}
  &\mathcal{F}^{(m-1)} = \{l:|\hat{x}_l^{(m-1)}|<\tau_1\}, \\
  &\mathcal{E}^{(m-1)} = \{(l,l')\in \mathcal{E}, 
    |\hat{x}_l^{(m-1)} - \hat{x}_{l'}^{(m-1)}|<\tau_2\} \ .
\end{split}
\end{equation}

Denote $x_{ll'} = x_l - x_{l'}$ and define $\boldsymbol{\xi} = (x_1, \cdots, x_r, x_{12}, \cdots, x_{1r}, \cdots, x_{(r-1)r})$. The $m$-th subproblem Eq.(\ref{eq:sub-problem}) can be reformulated as an equality-constrained convex optimization problem, 
\begin{equation}\label{eq:sub-eq-const}
  \min_{\boldsymbol{\xi}} \sum_{i=1}^n [(b_i - \ba_i^T\bx)^2 + \epsilon]^{\frac{1}{2}} 
    + \frac{\lambda_1}{\tau_1} \sum_{l \in \mathcal{F}^{(m-1)}} |x_l| 
    + \frac{\lambda_2}{\tau_2} \sum_{l<l': (l,l')\in \mathcal{E}^{(m-1)}} | x_{ll'}| 
    + \lambda_3 \sum_{l=1}^r x_l^2 ,
\end{equation}
subject to 
\begin{equation*}
  x_{ll'} = x_l - x_{l'},  \forall l<l':(l,l')\in \mathcal{E}^{(m-1)} .
\end{equation*}
For the equality-constrained problem Eq.(\ref{eq:sub-eq-const}), we employ the augmented Lagrange method to solve its equivalent unconstrained version iteratively with respect to $k$ for the $m$-th approximation.

\subsection{Alternating direction method of multipliers (ADMM)}

To apply ADMM in our constrained optimization problem Eq.(\ref{eq:sub-eq-const}), we separate Eq.(\ref{eq:sub-eq-const}) in two parts:
\begin{equation}\label{eq:sub-eq-xz}
\begin{split}
  &f(\bx) 
    = \sum_{i=1}^n [(b_i - \ba_i^T\bx)^2 + \epsilon]^{\frac{1}{2}} 
      + \frac{\lambda_1}{\tau_1} \sum_{l \in \mathcal{F}^{(m-1)}} |x_l|  
      + \lambda_3 \sum_{l =1}^r x_l^2, \\
  &g(\bx_{ll'}) 
    = \frac{\lambda_2}{\tau_2} \sum_{l<l': (l,l')\in \mathcal{E}^{(m-1)}} | x_{ll'}|,
\end{split}
\end{equation}
subject to
\begin{equation*}
  x_l - x_{l'} = x_{ll'}, \quad \forall l<l', (l,l') \in \mathcal{E}^{(m-1)}.  
\end{equation*}

With definition $\boldsymbol{\xi} = (\bx, \bx_{ll'})$, 
the augmented Lagrangian for Eq.(\ref{eq:sub-eq-xz}) is
\begin{equation}\label{eq:alm}
\begin{split}
  L_{\nu}^{(m)}(\boldsymbol{\xi}, \boldsymbol{\tau}) 
  = L_{\nu}^{(m)}(\bx, \bx_{ll'}, \boldsymbol{\tau}) 
  = & f(\bx) 
      + g(\bx_{ll'}) 
      + \sum_{l<l': (l,l')\in \mathcal{E}^{(m-1)}} \tau_{ll'}(x_l - x_{l'} - x_{ll'}) \\
    & + \frac{\nu}{2} \sum_{l<l': (l,l')\in \mathcal{E}^{(m-1)}} (x_l - x_{l'} - x_{ll'})^2 ,\\
\end{split}
\end{equation}
where $\tau_{ll'}$ and $\nu$ are the Lagrangian multipliers for the linear constraints and for the computational acceleration.

Minimizing Eq.(\ref{eq:alm}) over $ \boldsymbol{\xi} = (\bx, \bx_{ll'})$ yields $\widehat{\boldsymbol{\xi}}^{(m,k)} = (\widehat{\bx}^{(m, k)}, \widehat{\bx}_{ll'}^{(m, k)} ) $ for given values of $(\boldsymbol{\tau}_{ll'}^{k},\nu^{k})$. 
Note that as the iteration proceeds, $x_l - x_{l'}$ and $x_{ll'}$ becomes closer and closer, so we need $\nu$ to increase through the process. In particular, the ADMM updating rules are as follows

\begin{align}
 & \bx \text{--updating} \quad  & \widehat{\bx}^{(m,k+1)} 
 = \argmin_{\bx} L^{(m)}_{\nu} (\bx, \widehat{\bx}_{ll'}^{(m, k)}, \boldsymbol{\tau}^k), \\
 & \bx_{ll'} \text{--updating} \quad & \widehat{\bx}_{ll'}^{(m,k+1)} 
 = \argmin_{\bx_{ll'}} L^{(m)}_{\nu} ( \widehat{\bx}^{(m, k+1)},\bx_{ll'}, \boldsymbol{\tau}^k), \\
 & \boldsymbol{\tau}_{ll'} \text{--updating}\quad & \tau_{ll'}^{k+1} 
    = \tau_{ll'}^{k} + \nu^{k} 
      \left(\hat{x}_l^{(m,k+1)} - \hat{x}_{l'}^{(m,k+1)} - \hat{x}_{ll'}^{(m,k+1)} \right), \quad 
  \nu^{k+1} = \rho \nu^{k} .
\end{align}
Here $\rho$ is some constant chosen larger than $1$, which controls the acceleration of the algorithm. The details of the $\widehat{\boldsymbol{\xi}}^{(m,k)}$ minimization step can be found in the following two subsections.

\subsection{The $\bx$--updating}

\paragraph{Updating $\hat{x}_l^{(m,k)}$ in ADMM}

We first write the function of $\bx$ as $H(\bx)$.
\begin{equation*}
\begin{split}
    H(\bx) &\overset{\Delta}{=} L^{(m)}_{\nu} (\bx, \widehat{\bx}_{ll'}^{(m, k)}, \boldsymbol{\tau}^k) \\
        & = \underbrace{
        \sum_{i=1}^n [(b_i - \ba_i^T\bx)^2 + \epsilon]^{\frac{1}{2}} }_{\langle 1 \rangle}
        + \underbrace{
        \frac{\lambda_1}{\tau_1} \sum_{l\in \mathcal{F}^{(m-1)}}|x_l|}_{\langle 2 \rangle}
        +\underbrace{
        \lambda_3 \sum_{l=1}^r x_l^2}_{\langle 3 \rangle}\\
        &+\underbrace{
        \sum_{l<l': (l,l') \in \mathcal{E}^{(m-1)}} \tau_{ll'}^k(x_l - \hat{x}_{l'}^{(m,k-1)} -\hat{x}_{ll'}^{(m,k-1)})}_{\langle 4-1 \rangle}
        + \underbrace{
        \sum_{l>l': (l,l') \in \mathcal{E}^{(m-1)}} \tau_{ll'}^k(x_l - \hat{x}_{l'}^{(m,k)} -\hat{x}_{ll'}^{(m,k-1)})}_{\langle 4-2 \rangle}\\
        &+\underbrace{
        \frac{\nu^k}{2} \sum_{l<l': (l,l') \in \mathcal{E}^{(m-1)}} (x_l - \hat{x}_{l'}^{(m,k-1)} -\hat{x}_{ll'}^{(m,k-1)})^2}_{\langle 5-1 \rangle}
        + \underbrace{
        \frac{\nu^k}{2} \sum_{l>l': (l,l') \in \mathcal{E}^{(m-1)}} (x_l - \hat{x}_{l'}^{(m,k)} -\hat{x}_{ll'}^{(m,k-1)})^2}_{\langle 5-2 \rangle}\\
\end{split}
\end{equation*}

We calculate the derivative of $H(\bx)$ part by part. Set the derivatives of $\langle 1 \rangle + \langle 2 \rangle+ \langle 3 \rangle
+ \langle 4-1 \rangle+ \langle 4-2 \rangle
+ \langle 5-1 \rangle+ \langle 5-2 \rangle = 0$, we get an equation about $x_l$.

The $x_l$ terms:
\begin{equation*}
    \sum_{i=1}^n D_i^{-\frac{1}{2}} A_{il}^2x_l
    + 2\lambda_3 \cdot x_l 
    + \nu^{k} \sum_{l<l': (l,l')\in \mathcal{E}^{(m-1)}} 1 \cdot x_l
    + \nu^{k} \sum_{l>l': (l,l')\in \mathcal{E}^{(m-1)}} 1 \cdot x_l
\end{equation*}
where 
\begin{equation*}
    \nu^{k} \sum_{l<l': (l,l')\in \mathcal{E}^{(m-1)}} 1 \cdot x_l
    + \nu^{k} \sum_{l>l': (l,l')\in \mathcal{E}^{(m-1)}} 1 \cdot x_l 
  = \nu^k \cdot 
  \left\vert l':(l,l')\in \mathcal{E}^{(m-1)} \right\vert 
  \cdot x_l \
\end{equation*}
and $ \left\vert l':(l,l')\in \mathcal{E}^{(m-1)} \right\vert $
is the number of the elements in the set.

The constant terms:
\begin{equation*}
\begin{split}
    & -\sum_{i=1}^n D_i^{-\frac{1}{2}} 
    ( b_i A_{il} - (\ba_{i,-l} \widehat{\bx}_{-l}^{(m,k-1)})\cdot A_{il})
     + \Bigg \{ \begin{matrix}
        &\frac{\lambda_1}{\tau_1} 
        & \text{ if }|\hat{x}_l^{(m-1)}|<\tau_1 \text{ and }x_l>0\\
        &-\frac{\lambda_1}{\tau_1}
        & \text{ if }|\hat{x}_l^{(m-1)}|<\tau_1 \text{ and }x_l<0\\
        &0 & \text{otherwise}
    \end{matrix}\\
    &+  \sum_{(l,l')\in \mathcal{E}^{(m-1)}} \tau_{ll'}^k
     -  \nu^{k} \sum_{(l,l')\in \mathcal{E}^{(m-1)}} 
       (\hat{x}_{l'}^{(m,k-1)} + \hat{x}_{ll'}^{(m,k-1)})
\end{split}
\end{equation*}

Thus the updating of $\hat{x}_l^{(m,k)}$,
\begin{equation*}
\hat{x}_l^{(m,k)} = \alpha^{-1} \gamma ,
\end{equation*}
where 
\begin{equation*}
    \alpha = \sum_{i=1}^n D_i^{-\frac{1}{2}} A_{il}^2
    + 2\lambda_3  
    + \nu^k \left\vert l':(l,l') \in \mathcal{E}^{(m-1)}\right\vert ,
\end{equation*}
and 
\begin{equation*}
\gamma = 
    \begin{cases}
    \gamma^{*}, 
    & \text{ if } |\hat{x}_l^{(m-1)}| \geq \tau_1 \\
    \mathrm{ST}(\gamma^{*}, \frac{\lambda_1}{\tau_1}), 
    & \text{ if } |\hat{x}_l^{(m-1)}| < \tau_1
    \end{cases} \ .
\end{equation*}
ST is the soft threshold function, 
\begin{equation*}
\mathrm{ST}(x, \delta) = \sign(x) (|x| - \delta)_{+} 
= \begin{cases}
    x-\delta, 
    & \text{ if } b > \delta \\
    x+\delta, 
    & \text{ if } b <- \delta \\
    0, 
    & \text{ if } |b| \leq \delta 
    \end{cases} \ ,
\end{equation*}
and 
\begin{equation*}
  \gamma^{*} = 
    \sum_{i=1}^n D_i^{-\frac{1}{2}} c_{il} - \sum_{l<l': (l,l') \in \mathcal{E}^{(m-1)}} \tau^k_{ll'} 
     + \nu^k \sum_{l<l': (l,l') \in \mathcal{E}^{(m-1)}} (\hat{x}_{l'}^{(m,k-1)} 
    + \hat{x}_{ll'}^{(m,k-1)}) \ ,
\end{equation*}
where 
$ { D_i = (b_i-\ba_i^T \widehat{\bx}^{(m,k-1)})^2+\epsilon } $,  
$ { c_{il} = A_{il}\cdot (b_i - \ba_{i,-l}^T \widehat{\bx}_{-l}^{(m,k-1)}) } $, 
$\ba_{i,-l} $ is the $i$-th row vector of $\bA$ without the $l$-th column, 
and  $\widehat{\bx}_{(-l)}^{(m,k-1)}$ is the vector $\widehat{\bx}^{(m,k-1)}$ without the $l$-th component.

\subsection{ The $\bx_{ll'}$ \text{--updating} }

\paragraph{Updating $\hat{x}_{ll'}^{(m,k)}$ in ADMM}

Given $\hat{x}_{ll'}^{(m,k-1)}$, update $\hat{x}_{ll'}^{(m,k)} (1\leq l<l' \leq r)$ with $\hat{x}_l^{(m,k)}$ already updated and fixed.
\begin{equation*}
\begin{split}
Q(\bx_{ll'})
& \overset{\Delta}{ = } 
  L^{(m)}_{\nu} ( \widehat{\bx}^{(m, k)},\bx_{ll'}, \boldsymbol{\tau}^k) \\
      & = \underbrace{
      \frac{\lambda_2}{\tau_2} \sum_{l<l': (l,l')\in \mathcal{E}^{(m-1)}} |x_{ll'}|}_{(\RNum{1})} 
      + \underbrace{
      \tau_{ll'}^{k} \sum_{l<l': (l,l')\in \mathcal{E}^{(m-1)}} (\hat{x}_l^{(m,k)} - \hat{x}_{l'}^{(m,k)} - x_{ll'})}_{(\RNum{2})}\\
      & + \underbrace{
      \sum_{l<l': (l,l')\in \mathcal{E}^{(m-1)}} \frac{\nu^k}{2} (\hat{x}_l^{(m,k)} - \hat{x}_{l'}^{(m,k)} - x_{ll'})^2}_{(\rom{3})} \ .
\end{split} 
\end{equation*}

\begin{flalign*}
\text{We calculate} \frac{\partial Q(\bx_{ll'}) }{\partial x_{ll'}} \text{ by part. Setting the derivative equals to zero, we have that}  && 
\end{flalign*}

\begin{equation*}
    0 = \nu^k x_{ll'} - \tau_{ll'}^k 
        - \nu^k(\hat{x}_l^{(m,k)} - \hat{x}_{l'}^{(m,k)}) + 
        \begin{cases}
        \frac{\lambda_2}{\tau_2}, & \text{ if } x_{ll'}>0 \\
        - \frac{\lambda_2}{\tau_2}, &\text{ if } x_{ll'}<0 
        \end{cases} \ .
\end{equation*}
Then 
\begin{equation}\label{eq:betajj-update}
\hat{x}_{ll'}^{(m,k)} = 
    \begin{cases}
    \frac{1}{\nu^k} 
     \mathrm{ST}(\tau_{ll'}^k + \nu^k(\hat{x}_l^{(m,k)} - \hat{x}_{l'}^{(m,k)}), \frac{\lambda_2}{\tau_2}), 
      & \text{ if }(l,l') \in \mathcal{E}^{(m-1)} \\
    \hat{x}_{ll'}^{(m-1)}, 
    & \text{ if }(l,l') \not\in \mathcal{E}^{(m-1)}
    \end{cases} \ .
\end{equation}

\section{Extension to Non-negative GRMF (N-GRMF)}
\subsection{Formulation for N-GRMF}
In this section we show that our GRMF model can be easy extended to the robust non-negative MF with grouping effect (N-GRMF). The problem is formulated as follows
\begin{equation}\label{eq:N-GRMF-obj}
  \min_{\bU \in \mathbb{R}^{d\times r}, \bV \in \mathbb{R}^{n\times r} } f(\bU, \bV) = 
  \|\bY - \bU\bV^{T}\|_{\ell_1} + \mathcal{R}(\bU) + \mathcal{R}(\bV)\ , 
\end{equation}
where $\mathcal{R}(\bU)$ and $\mathcal{R}(\bV)$
are two regularizers corresponding to $\bU$ and $\bV$, given by
\begin{equation*}
\begin{split}
\mathcal{R}(\bU) = &
  \sum_{i=1}^{d} \lambda_{1} \cP_1(\bu_i)
  + \sum_{i=1}^{d} \lambda_{2} \cP_2(\bu_i)
  + \sum_{i=1}^{d} \lambda_{3} \Tilde{\cP}_3(\bu_i), 
  \text{ and } \\
  \mathcal{R}(\bV) = &
  \sum_{j=1}^{n} \lambda_{1} \cP_1(\bv_j)
  + \sum_{j=1}^{n} \lambda_{2} \cP_2(\bv_j)
  + \sum_{j=1}^{n} \lambda_{3} \Tilde{\cP}_3(\bv_j)\ . 
\end{split}
\end{equation*}

Here $\cP_1(\cdot)$ and $\cP_2(\cdot)$ are the same as in GRMF, while $\Tilde{\cP}_3(\cdot)$ is slightly different from ${\cP}_3(\cdot)$, which takes the following form, 
\begin{equation}
\cP_1(\bx) = \sum_{l=1}^{r} \min\left(\frac{|x_l|}{\tau_1}, 1\right), \ 
\cP_2(\bx) = 
 \sum_{l<l': (l, l')\in \cE} \min\left(\frac{|x_{l}-x_{l'}|}{\tau_2}, 1\right), \ 
\Tilde{\cP}_3(\bx) = \sum_{l=1}^{r} (\min(x_l,0))^2\
\end{equation}

We still adopt the $\ell_1$-loss to attain the robustness and solve the problem by fixing $\bU$ or $\bV$ and updating the other one.

\subsection{Algorithms for N-GRMF}
The solution for N-GRMF is very similar to that of GRMF. Thus, we will not walk through the whole solution here again. 
Instead, we only detail what differs in this model.

\subsubsection{The DC algorithm}
Note that in this model $\Tilde{\cP}_3(\bx)$ needs to be decomposed by DC just as $\cP_1(\bx)$ and $\cP_2(\bx)$ do. In particular, $S(\bx)$ can be decomposed as follows, 
\begin{align*}
&S_1(\bx) = \sum_{i=1}^n [(b_i - \ba_i^T\bx)^2 + \epsilon]^{\frac{1}{2}} 
+ \lambda_1 \sum_{l=1}^{r} \frac{|x_l|}{\tau_1} 
+ \lambda_2 \sum_{l<l':(l,l')\in \mathcal{E}} \frac{|x_l-x_{l'}|}{\tau_2} + \lambda_3 \sum_{l=1}^{r} x_l^2 \\
&S_2(\bx) = \lambda_1 \sum_{l=1}^{r} \left(\frac{|x_l|}{\tau_1} - 1\right)_{+} 
    + \lambda_2 \sum_{l<l':(l,l')\in \mathcal{E}} \left(\frac{|x_l-x_{l'}|}{\tau_2}-1\right)_{+}
    + \lambda_3 \sum_{l=1}^r [(x_l)_{+}]^2 \ .
\end{align*}
For the $m$-th minimization, $S_{2}(\bx)$ is linearized and the subproblem becomes as follows, 
\begin{equation}\label{eq:m-beta-nn}
S^{(m)}(\bx) = 
  \sum_{i=1}^n [(b_i - \ba_i^T\bx)^2 + \epsilon]^{\frac{1}{2}} 
  + \frac{\lambda_1}{\tau_1}\sum_{l \in \mathcal{F}^{(m-1)}} |x_l| 
+ \frac{\lambda_2}{\tau_2}\sum_{l<l': (l,l') \in \mathcal{E}^{(m-1)}} |x_l - x_{l'}| 
  + \lambda_3 \sum_{l \in \mathcal{N}^{(m-1)}} x_l^2,
\end{equation}
where
\begin{align*}\label{eq:EF-update-nonnegative}
  &\mathcal{F}^{(m-1)} = \left\{l:|\hat{x}_l^{(m-1)}|<\tau_1 \right\},\\
  &\mathcal{E}^{(m-1)} = \left\{(l,l')\in \mathcal{E}, 
    |\hat{x}_l^{(m-1)} - \hat{x}_{l'}^{(m-1)}|<\tau_2 \right\} , \\
  &\mathcal{N}^{(m-1)} = \left\{l:\hat{x}_l^{(m-1)}<0 \right\} \ .
\end{align*}

Denote $x_{ll'} = x_l - x_{l'}$, 
and define $ { {\boldsymbol{\xi}} = (x_1, \cdots, x_r, x_{12}, \cdots, x_{1r}, \cdots, x_{(r-1)r}) }$. The $m$-th subproblem~\eqref{eq:m-beta-nn} can be reformulated as an equality-constrained convex optimization problem, 
\begin{equation}\label{eq:constraint-nn}
S^{(m)}(\boldsymbol{\xi}) 
  = \sum_{i=1}^n [(b_i - \ba_i^T\bx)^2 + \epsilon]^{\frac{1}{2}} 
    + \frac{\lambda_1}{\tau_1} \sum_{l \in \mathcal{F}^{(m-1)}} |x_l| 
    + \frac{\lambda_2}{\tau_2} \sum_{l<l': (l,l')\in \mathcal{E}^{(m-1)}} | x_{ll'}| 
    + \lambda_3 \sum_{l \in \mathcal{N}^{(m-1)}} x_l^2,
\end{equation}
subject to $x_{ll'} = x_l - x_{l'}, \quad  \forall l<l':(l,l')\in \mathcal{E}^{(m-1)}$.

\subsubsection{ADMM}
The only thing different for N-GRMF in this step is the updating rule of $\hat{x}_l^{(m,k)}$, since $\cP_3$ does not involve other variables but $x_l$. In particular, when updating by $\hat{x}_l^{(m,k)} = \alpha^{-1} \gamma$, $\alpha$ is a little different compared to the solution of GRMF and $\gamma$ stays the same: it is formulated as follows,
\begin{equation}
    \alpha = \sum_{i=1}^{n} D_i^{-\frac{1}{2}} A_{il}^2
        + 2\lambda_3 I_{\{\hat{x}_l^{(m-1)}<0\}}
        + \nu^k \left\vert l':(l,l') \in \mathcal{E}^{(m-1)}\right\vert,
\end{equation}
where $I_{\{\hat{x}_l^{(m-1)}<0\}} = 1$ when $\hat{x}_l^{(m-1)}<0$ and $I_{\{\hat{x}_l^{(m-1)}<0\}} = 0$ otherwise.

\section{More experiments}
\subsection{Additions to the main results}

\begin{figure}[!htb]
\includegraphics[width=1\textwidth]{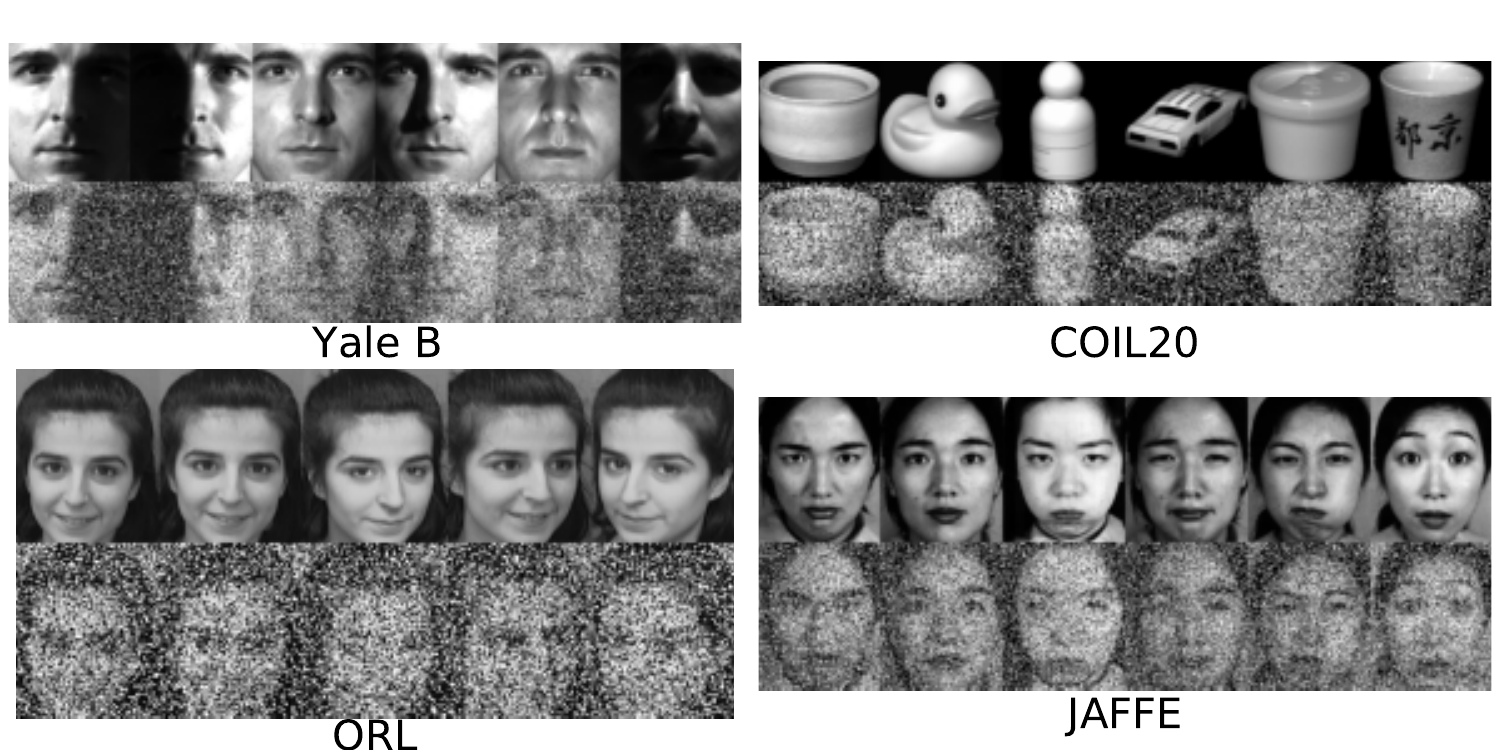} 
\caption{The original images and the corrupted versions. the first row of images is the original version, and the second row of images is the corrupted version with $50\%$ salt and pepper noise. } \label{fig:main_faces1}
\end{figure}

\autoref{fig:main_faces1} is a demonstration of the four datasets we used in the experiments. 
Each dataset consists of a large number of people or objects in different angles/expressions/light conditions. 
In this demonstration, the first row of images is the original version, and the second row of images is the corrupted version with $50\%$ salt and pepper noise.
In the experiment, we use each image as the input data, and then decompose it into two smaller matrices whose product constitutes the reconstructed image. We compare the reconstruction error between the original image and the reconstructed one.

\begin{figure*}[!htb]
{\includegraphics[width=1\textwidth]{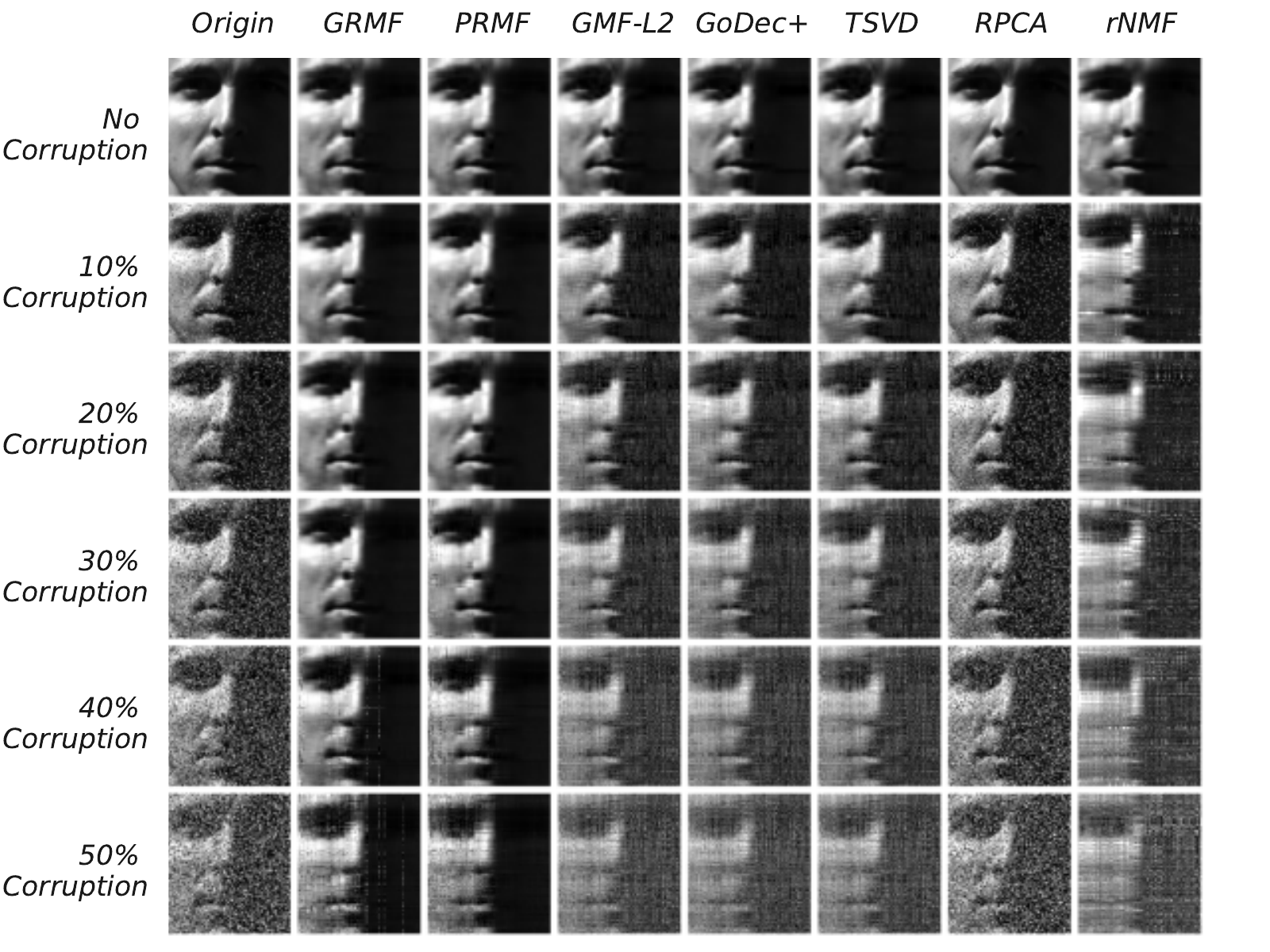}}
\caption{An example of image recovery with different MF methods under different corruption ratios.} 
\label{fig:main_face2} 
\end{figure*}

\autoref{fig:main_face2} shows one example of image recovery. The first column consists of input images with different corruption ratios. Each row represents the output images recovered by different methods. 
Among all methods, GRMF and PRMF demonstrate great recovery ability, especially when the corruption ratio is as high as $50\%$, where the output images look very similar to the original undamaged ones.
Unlike these two, RPCA recovers the input image to the greatest extent, but it does not show robustness when the input image is corrupted with noise.

As for the time cost,~\autoref{tab:main-result-time} demonstrates the average running time of different algorithms on the four datasets, including both the original version and the $50\%$ corrupted version. 
Our algorithm is more time consuming compared to the benchmarks because we apply an algorithm which consists of multi-layer nested loops. 
To be specific, we update each row (column) of $\bU(\bV)$ independently, and each subproblem is solved with two nested loops combining DC and coordinate-wise ADMM. 
This structure is necessary in our approach because of the $\ell_0$-surrogate regularization and the grouping effect we are pursuing.
The $\ell_0$-surrogate penalties are non-convex and thus can only be solved after decomposition into a difference of two convex functions. 
In addition, to pursue the grouping effect, we need to compare every pair of values in the output vector. 
Thus the resulting optimization problem could be solved by ADMM. 
In spite of that, 
the promoted performance of GRMF makes it worth being studied.
The grouping effect of GRMF helps denoising the factorization and makes it less sensitive to outliers and corruption, which is demonstrated by the great recovery ability in the experiments of $50\%$ corrupted images. Therefore, to enjoy the robustness and accuracy of GRMF, one significant future development of GRMF is to accelerate the training process.

\begin{table*}[tb]
\centering
\caption{Comparison of running time on the four datasets. The average time cost (seconds) per image is reported with the standard deviation in the parentheses. T-SVD stands for the Truncated SVD.}

\label{tab:main-result-time}

\setlength\tabcolsep{2pt}

\resizebox{\textwidth}{!}{  
\begin{tabular}{ccccccccc}

\toprule

\multirow{2}{*}{Datasets} & \multicolumn{2}{c}{Yale B} & \multicolumn{2}{c}{COIL} & \multicolumn{2}{c}{ORL} & \multicolumn{2}{c}{JAFFE}\\

\cmidrule(r){2-3} \cmidrule(r){4-5} \cmidrule(r){6-7}  \cmidrule(r){8-9}

&  Origin  &  Corrupted
&  Origin  &  Corrupted
&  Origin  &  Corrupted
&  Origin  &  Corrupted \\

\midrule

GRMF &  
155.7$\pm$(10.3) & 334.6$\pm$(75.4) & 188.4$\pm$(66.2) & 480.7$\pm$(409.2) & 85.1$\pm$(11.7) & 159.2$\pm$(16.0) & 177.4$\pm$(10.5) & 353.4$\pm$(40.9)  \\

PRMF & 
3.3$\pm$(0.1) & 3.1$\pm$(0.1) & 8.8$\pm$(20.3) & 2.3$\pm$(0.6) & 1.4$\pm$(0.1) & 1.3$\pm$(0.0) & 3.2$\pm$(0.1) & 3.1$\pm$(0.2)  \\

GMF-L2 & 
27.2$\pm$(6.6) & 20.6$\pm$(0.1) & 42.0$\pm$(43.6) & 22.5$\pm$(104.0) & 21.3$\pm$(5.3) & 9.0$\pm$(1.1) & 22.5$\pm$(4.6) & 18.1$\pm$(1.8)  \\

GoDec+ & 
0.1$\pm$(0.0) & 0.0$\pm$(0.0) & 0.1$\pm$(0.3) & 0.0$\pm$(0.0) & 0.0$\pm$(0.0) & 0.0$\pm$(0.0) & 0.0$\pm$(0.0) & 0.0$\pm$(0.0)  \\

T-SVD &
0.0$\pm$(0.0) & 0.0$\pm$(0.0) & 0.0$\pm$(0.0) & 0.0$\pm$(0.0) & 0.0$\pm$(0.0) & 0.0$\pm$(0.0) & 0.0$\pm$(0.0) & 0.0$\pm$(0.0)  \\

RPCA & 
5.4$\pm$(0.1) & 0.6$\pm$(0.1) & 6.5$\pm$(12.9 & 3.5$\pm$(0.4) & 1.6$\pm$(0.1) & 1.8$\pm$(0.1) & 4.7$\pm$(0.1) & 0.6$\pm$(0.1)  \\

rNMF & 
0.5$\pm$(0.5) & 5.1$\pm$(0.2) & 6.5$\pm$(10.9) & 3.0$\pm$(0.5) & 1.7$\pm$(0.1) & 1.8$\pm$(0.0) & 4.5$\pm$(0.2) & 4.5$\pm$(0.1)  \\

\bottomrule
\end{tabular}
}
\end{table*}




\subsection{Non-negative extension results}
\begin{figure*}[tb]
\subfloat[\centering RMAE of Yale B ]{\includegraphics[width=0.45\textwidth]{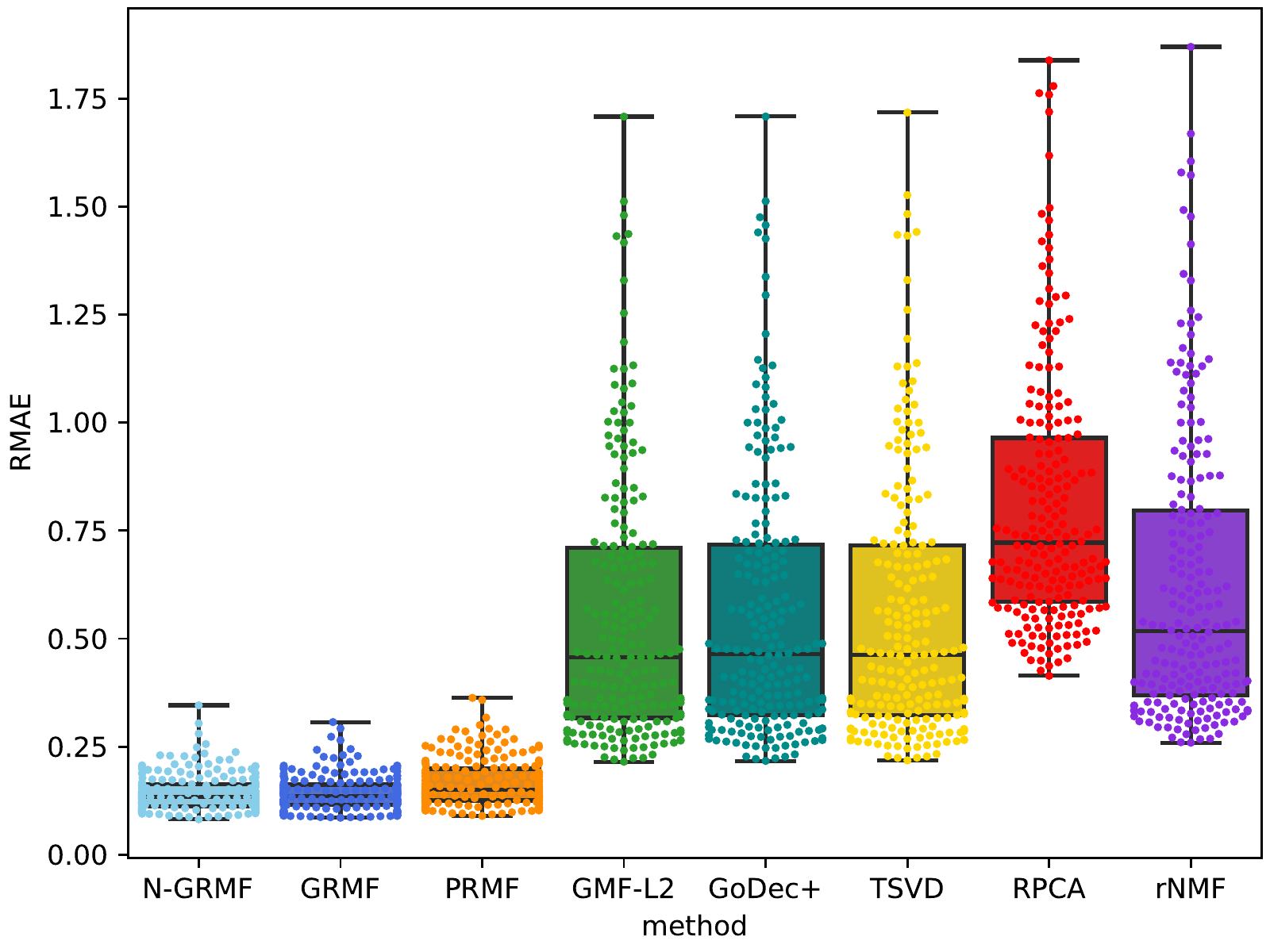} \label{fig:Yale B} }%
\subfloat[\centering RMAE of COIL ]{\includegraphics[width=0.45\textwidth]{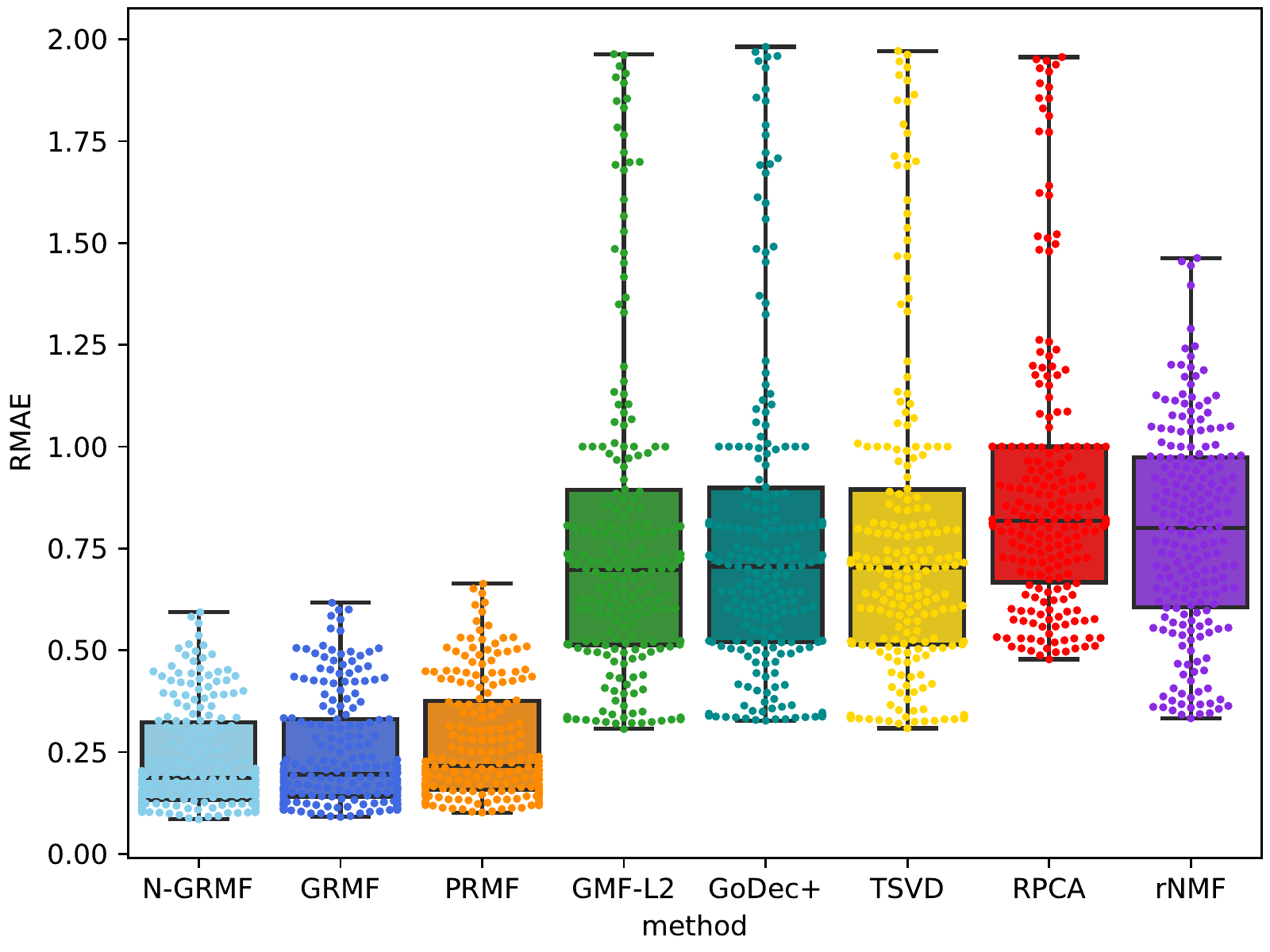} \label{fig:COIL} }%
\\
\subfloat[\centering RMAE of ORL ]{\includegraphics[width=0.45\textwidth]{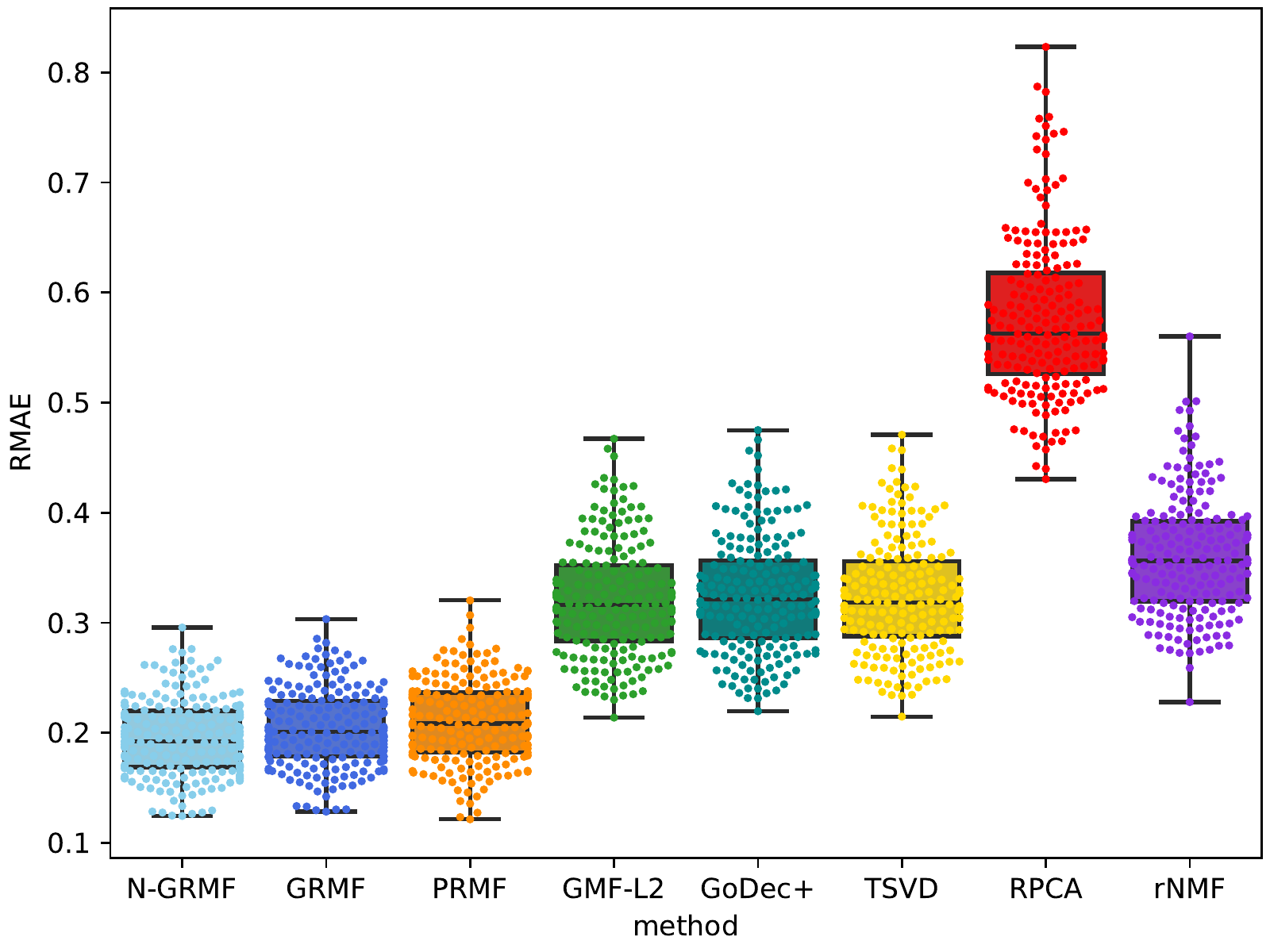} \label{fig:ORL} }%
\quad
\subfloat[\centering RMAE of JAFFE ]{\includegraphics[width=0.45\textwidth]{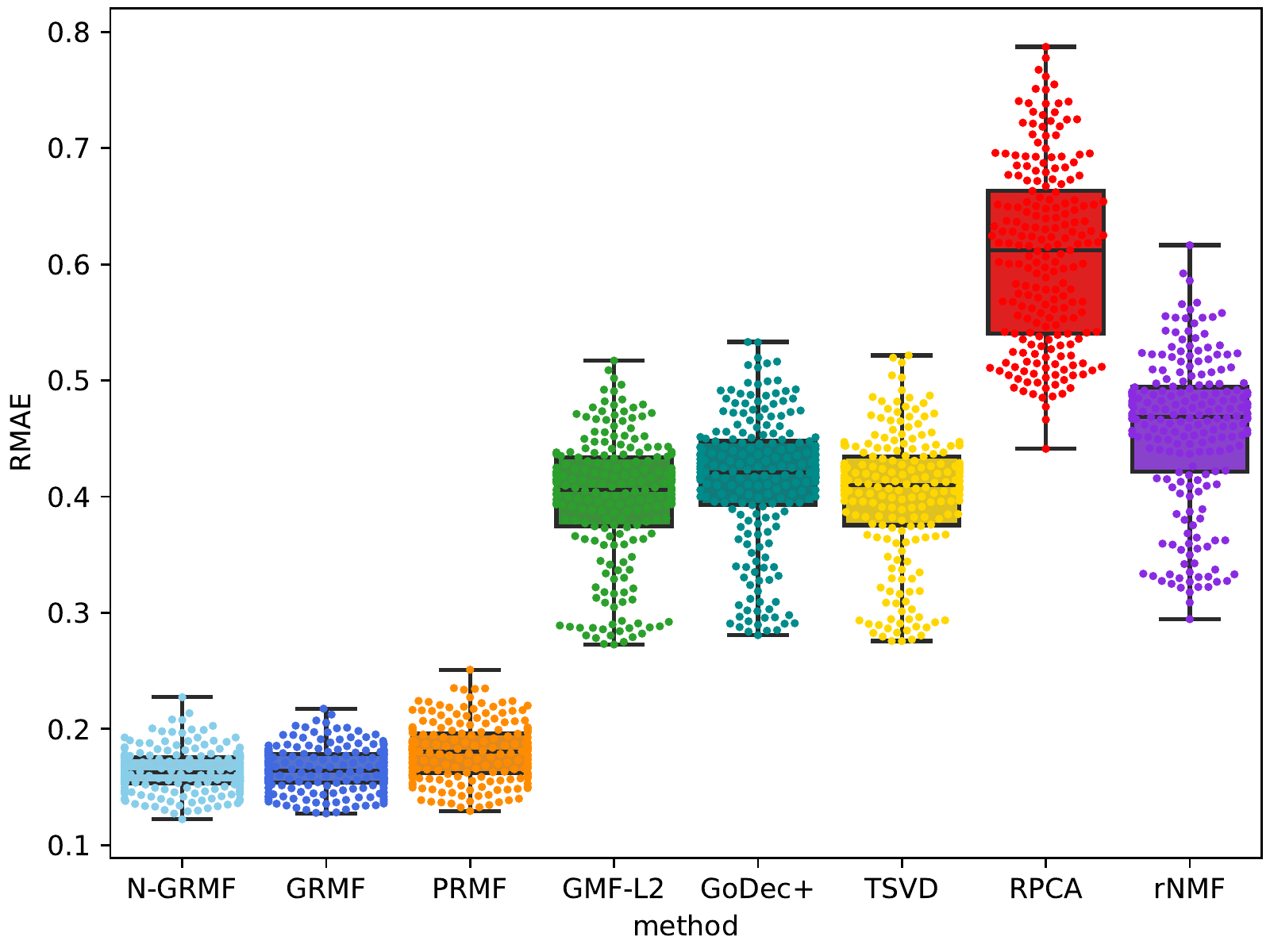} \label{fig:JAFFE} }%
\caption{Comparison of the distribution of relative mean absolute error(RMAE) among non-negative GRMF, regular GRMF, and all the benchmarks on four datasets with $50\%$ corruption ratio. N-GRMF stands for non-negative GRMF, and T-SVD stands for the truncated SVD.}
\label{fig:boxplot-RMAE}
\end{figure*}

We extend our GRMF model to a non-negative variant (N-GRMF) and conduct matrix factorization using the same four datasets in this section. The algorithm of N-GRMF is covered in {\color{blue}Appendix B}. 
Then~\autoref{tab:main_table2} demonstrates a comparison of the relative mean absolute error (RMAE) between non-negative GRMF, regular GRMF, and PRMF on four datasets with both its original version and the $50\%$ corrupted version. 
The RMAE of GRMF and N-GRMF remains at the same level when doing factorization with respect to the origin image. 
Both GRMF and N-GRMF have a minor improvement on the corrupted image recovery. 
~\autoref{fig:boxplot-RMAE} illustrates the comparison of the reconstruction error between N-GRMF, regular GRMF, and all the benchmarks on four datasets with $50\%$ corruption ratio. 
The distribution of the reconstruction error for each image is represented by the box plot and scatter plot, which shows the maximum, upper quantile, mean, standard deviation, lower quantile, minimum. Besides, every error point is plotted over it. 
As illustrated in~\autoref{tab:main_table2} and ~\autoref{fig:boxplot-RMAE}, the error of N-GRMF has a minor decrease compared with that of the regular GRMF without the non-negative constraint, when the input data is corrupted, implying an enhanced robustness from GRMF to N-GRMF.

\begin{table*}[tb]
\centering
\caption{Comparison of Relative MAE among N-GRMF, regular GRMF, and PRMF on four datasets with $50\%$ corruption ratio. The mean RMAE is reported with the standard deviation in the parentheses. N-GRMF stands for non-negative GRMF.}

\label{tab:main_table2}
\setlength\tabcolsep{2pt}

\resizebox{\textwidth}{!}{  

\begin{tabular}{ccccccccc}

\toprule

\multirow{2}{*}{Datasets} & \multicolumn{2}{c}{Yale B} & \multicolumn{2}{c}{COIL} & \multicolumn{2}{c}{ORL} & \multicolumn{2}{c}{JAFFE}\\

\cmidrule(r){2-3} \cmidrule(r){4-5} \cmidrule(r){6-7}  \cmidrule(r){8-9}

&  Origin  &  Corrupted
&  Origin  &  Corrupted
&  Origin  &  Corrupted
&  Origin  &  Corrupted \\

\midrule
N-GRMF & 
\textbf{0.093}$\pm$(0.022) & \textbf{0.143}$\pm$(0.042) & \textbf{0.123}$\pm$(0.066) & \textbf{0.234}$\pm$(0.126) & \textbf{0.105}$\pm$(0.020) & \textbf{0.196}$\pm$(0.035) & \textbf{0.121}$\pm$(0.013) & \textbf{0.164}$\pm$(0.018) \\

GRMF &  
\textbf{0.093}$\pm$(0.022) & \textbf{0.143}$\pm$(0.041) & \textbf{0.123}$\pm$(0.066) & 0.245$\pm$(0.135) & \textbf{0.105}$\pm$(0.020) & 0.204$\pm$(0.036) & \textbf{0.121}$\pm$(0.013) & 0.165$\pm$(0.018) \\

PRMF & 
0.095$\pm$(0.023) & 0.154$\pm$(0.047) & 0.127$\pm$(0.070) & 0.273$\pm$(0.142) & 0.107$\pm$(0.020) & 0.107$\pm$(0.037) & 0.125$\pm$(0.014) & 0.182$\pm$(0.025)  \\

\bottomrule

\end{tabular}
}
\end{table*}

\end{document}